\documentclass[preprint,
prx,amsmath,amssymb
]{revtex4-2}

\usepackage[margin=1in]{geometry}
\usepackage{graphicx}
\usepackage{dcolumn}
\usepackage{bm}
\usepackage{setspace}
\usepackage{lineno}
\usepackage{amsmath}
\usepackage{amssymb}
\newcommand{\vv}[1]{\textbf{#1}}
\usepackage[T1]{fontenc}
\usepackage[utf8]{inputenc}
\usepackage[english]{babel}
\usepackage{hyperref}

\begin{document}
\title{Combining Machine Learning with Knowledge-Based Modeling for Scalable Forecasting and Subgrid-Scale Closure of Large, Complex, Spatiotemporal Systems}
\date{\today}
\author{Alexander Wikner}
\affiliation{Department of Physics and Institute for Research in Electronics and Applied Physics, University of Maryland, College Park, MD, 20740}
\author{Jaideep Pathak}
\affiliation{Department of Physics and Institute for Research in Electronics and Applied Physics, University of Maryland, College Park, MD, 20740}
\author{Brian Hunt}
\affiliation{Institute for Physical Science and Technology, University of Maryland, College Park, MD 20740}
\affiliation{Department of Mathematics, University of Maryland, College Park, MD 20740}
\author{Michelle Girvan}
\affiliation{Department of Physics and Institute for Research in Electronics and Applied Physics, University of Maryland, College Park, MD, 20740}
\affiliation{Institute for Physical Science and Technology, University of Maryland, College Park, MD 20740}
\author{Troy Arcomano}
\affiliation{Department of Atmospheric Sciences, Texas A\&M University, College Station, TX 77843}
\author{Istvan Szunyogh}
\affiliation{Department of Atmospheric Sciences, Texas A\&M University, College Station, TX 77843}
\author{Andrew Pomerance}
\affiliation{Potomac Research LLC, Alexandria, VA 22311}
\author{Edward Ott}
\affiliation{Department of Physics and Institute for Research in Electronics and Applied Physics, University of Maryland, College Park, MD, 20740}
\affiliation{Department of Electrical and Computer Engineering, University of Maryland, College Park, MD 20740}
\begin{abstract}
We consider the commonly encountered situation (e.g., in weather forecasting) where the goal is to predict the time evolution of a large, spatiotemporally chaotic dynamical system when we have access to both time series data of previous system states and an imperfect model of the full system dynamics. Specifically, we attempt to utilize machine learning as the essential tool for integrating the use of past data into predictions. In order to facilitate scalability to the common scenario of interest where the spatiotemporally chaotic system is very large and complex, we propose combining two approaches:(i) a parallel machine learning prediction scheme; and (ii) a hybrid technique, for a composite prediction system composed of a knowledge-based component and a machine-learning-based component. We demonstrate that not only can this method combining (i) and (ii) be scaled to give excellent performance for very large systems, but also that the length of time series data needed to train our multiple, parallel machine learning components is dramatically less than that necessary without parallelization. Furthermore, considering cases where computational realization of the knowledge-based component does not resolve subgrid-scale processes, our scheme is able to use training data to incorporate the effect of the unresolved short-scale dynamics upon the resolved longer-scale dynamics (``subgrid-scale closure'').
\end{abstract}

\maketitle

\section{Introduction}\label{sec:intro}
In recent years, machine learning techniques have been used to solve a number of complex modeling problems ranging from effective translation between hundreds of different human languages ~\cite{wu_googles_2016} to predicting the bioactivity of small molecules for drug discovery ~\cite{wallach_atomnet:_2015}. Typically, the most impressive results have been obtained using artificial neural networks with many hidden neural states ~\cite{lecun_deep_2015}. These hidden layers form a ``black box'' model where internal parameters are trained given a set of measured training data but, after which, only the final model output is observed. This formulation using measured training data contrasts with how models used for forecasting physical spatiotemporally chaotic processes are formulated, which is typically done using the available scientific knowledge of the underlying mechanisms that govern the system's evolution. For example, in the case of forecasting weather, this knowledge includes the Navier-Stokes equations, the first law of thermodynamics, the ideal gas law, and (see Sec.~\ref{sec:multiscale}) simplified representations of physics at the unresolved spatial scales~\cite{bauer_quiet_2015}. 


In this paper, focusing on the key issues of scalability and unresolved subgrid physics, we consider the general problem of forecasting a vary large and complex spatiotemporally chaotic system where we have access to both past time series of measurements of the system state evolution and to an imperfect knowledge-based prediction model. We present a method for combining machine learning prediction with imperfect knowledge-based forecasting that is scalable to large systems with the aim that the resulting combined prediction system can be significantly more accurate and efficient than either a pure knowledge-based prediction or a pure machine-learning-based prediction. A main source of difficulty for scalability of the machine learning is that the dimension of the state of the systems we are interested in can be extremely large. For example, in state-of-the-art global numerical weather models the state dimension (number of variables at all grid points) can be on the order of $10^9$. Thus, both the machine learning input (the current atmospheric state) and output (the predicted atmospheric state) have this dimensionality. (In contrast to the description of some machine learning techniques as ``deep'', one might refer to the situations we address as ``wide''.) The prediction method that we propose for such large complex systems builds on the previous work on parallelizable machine learning prediction~\cite{parallelreservoir2018} and hybridization of knowledge-based modeling with machine learning ~\cite{hybridreservoir2018}. We call our technique Combined Hybrid/Parallel Prediction (CHyPP, pronounced ``chip'').
Although the general method we propose is applicable to different kinds of machine learning, the numerical examples presented in this paper use a machine learning method known as reservoir computing ~\cite{herbert2001echo,Maass_LSM_2002}. Jaeger and Haas ~\cite{jaeger_harnessing_2004} described the effectiveness of reservoir computing for predicting low-dimensional chaotic systems. Research surrounding this technique has since expanded ~\cite{lu_reservoir_2017, inubushi_reservoir_2017}, and it has recently been shown that reservoir computing using recurrent neural networks can produce similar quality predictions for chaotic systems to those of other recurrent architectures, such as LSTM's ~\cite{LSTM_1997} and GRU's ~\cite{cho_learning_2014}, while often requiring much less computational time to train~\cite{vlachas_forecasting_2019}. Reservoir computing techniques can additionally be extended to physical implementations using, e.g., photonics ~\cite{vinckier_high-performance_2015} and Field Programable Gate Arrays (FPGA's)~\cite{penkovsky_efficient_2018, tanaka_recent_2019}.

The rest of the paper is organized as follows. In Sec.~\ref{sec:reservoir}, we first review a simple version of reservoir computing ~\cite{herbert2001echo,Maass_LSM_2002} and discuss its shortcomings for forecasting high-dimensional spatiotemporal chaos. We next describe the hybrid reservoir prediction technique (Refs. ~\cite{hybridreservoir2018} and ~\cite{wan2018data}), as well as previous work on how machine learning can be parallelized for prediction of spatiotemporal systems ~\cite{parallelreservoir2018}. We then present our proposed CHyPP architecture combining the two. In Sec.~\ref{sec:results}, we demonstrate how the CHyPP methodology improves on each of the component prediction methods. For these demonstrations we use the paradigmatic example of the Kuramoto-Sivashinky model as our test model of the spatiotemporally chaotic system that we aim to predict. We highlight the scalability of the proposed method to very large systems as well as its efficient use of training data, which we view as the crucial issues for the general class of applications in which we are interested. In Sec.~\ref{sec:multiscale}, we consider a situation with multiple time and space scales and show by numerical simulation tests, that CHyPP can, through its use of data, effectively account for unknown subgrid scale processes. The main conclusion of this paper is that our CHyPP methodology provides an extremely promising framework, potentially facilitating significant advances in the forecasting of large, complex, spatiotemporally chaotic systems. We believe that, in addition to weather, the method that we propose may potentially be applicable to a host of important areas, enabling currently unattainable capabilities. Some speculative examples of potential applications are forecasting of ocean conditions, forecasting conditions in the solar wind, magnetosphere and ionosphere (also known as 'space weather', important for its impact on Earth-orbiting satellites, GPS accuracy, and high frequency communications), forecasting the evolution of forest fires and their response to mitigating interventions, forecasting the responses of ecological systems to climate change, analysis of neuronal activity, etc.

\section{Reservoir Computing Architecture for CHyPP}\label{sec:reservoir}

\subsection{A Simple Machine Learning Predictor}
\begin{figure}[h!]
    \centering
    \includegraphics[width = \textwidth]{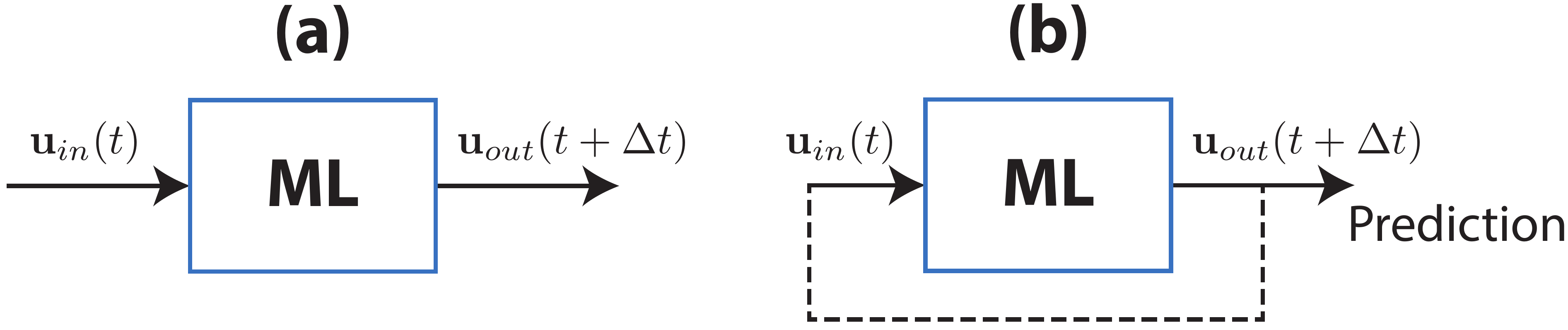}
    \caption{Machine learning prediction device (a) open-loop training phase and (b) closed-loop prediction phase.}
    \label{fig:basicML}
\end{figure}

To begin, we initially consider the goal of a generic machine learning system for time-series prediction of an unknown dynamical system evolving on an attractor of that system. Later, we will consider that the machine learning system is a reservoir computer and that the unknown chaotic system is not completely unknown, and we will try to make use of that knowledge. Given a finite duration time series of the unknown system state's evolution up to a certain time $t_0$, where the state at each time is represented by the $K$-dimensional vector $\vv{u}(t) = [u_1(t), u_2(t),\dots,u_K(t)]^T$, our goal is to predict the subsequent evolution of the state. As illustrated in Fig.~\ref{fig:basicML}(a), in the initial training phase, at each time $t=n\Delta t\leq t_0$, $\vv{u}(t)$ is input to the machine learning system ($\mathbf{u}_{in}(t) = \vv{u}(t)$), which is trained to output a time $\Delta t$ prediction of the dynamical system state $\vv{u}(t + \Delta t)$ ($\vv{u}_{out}(t+\Delta t) \simeq \vv{u}(t+\Delta t)$). We refer to the just-described input-output configuration as the ``open-loop'' system (Fig.~\ref{fig:basicML}a). To ensure an accurate representation of the true dynamics with a reservoir of limited size, $\Delta t$ is typically short compared to natural time scales (such as the correlation time or the ``Lyapunov time'') present in the unknown dynamical system whose state is to be predicted. Once trained, the machine learning system can be run in a ``closed-loop'' feedback configuration (Fig.~\ref{fig:basicML}(b)) to autonomously generate predictions over a finite duration of time. That is, with $t_0$ representing the time at the end of the training data, we replace the former input from the training data by the output by inserting an output-to-input feedback loop, shown by the dashed line in Fig.~\ref{fig:basicML}(b). Then, when $\vv{u}_{in}(t_0) = \vv{u}(t_0)$ is the input, the reservoir computer produces an output prediction for $\vv{u}(t_0+\Delta t)$, which we refer to as $\tilde{\vv{u}}(t_0+\Delta t) = \vv{u}_{out}(t_0+\Delta t)$. When this predicted state is then used as the input ($\vv{u}_{in}(t_0+\Delta t) = \tilde{\vv{u}}(t_0+\Delta t)$), the reservoir computer produces an output prediction for $\vv{u}(t_0+2\Delta t)$, denoted $\tilde{\vv{u}}(t_0+2\Delta t)$ ($\vv{u}_{out}(t_0+2\Delta t) = \tilde{\vv{u}}(t_0+2\Delta t)$). This process is then iterated to produce predictions of the system state at $t=t_0+m\Delta t$ for $m = 1,2,3,\dots$ (Fig.~\ref{fig:basicML}b). 

In the rest of this section, we first present background from previous work (Secs. ~\ref{sec:reservoir-basic}-~\ref{sec:parallelization}), then introduce our CHyPP method for combining a knowledge-based model with reservoir-based machine learning to form a scalable system concept suitable for state prediction of very large, spatiotemporally chaotic systems. Specifically in  Sec.~\ref{sec:reservoir-basic}, we review a basic reservoir computing setup based on the methods of~\cite{herbert2001echo,Maass_LSM_2002} along with the proposal for its use as a predictor carried out in Ref.~\cite{jaeger_harnessing_2004}. In Sec.~\ref{sec:hybrid}, we build upon the simple setup of Sec~\ref{sec:reservoir-basic} and describe the methodology from Ref.~\cite{parallelreservoir2018} for hybrid forecasting of the dynamical system using a single reservoir computing network and an imperfect model. In Sec.~\ref{sec:parallelization}, the reservoir computing forecasting technique of Sec.~\ref{sec:reservoir-basic} is extended via parallelization of the machine learning with multiple parallel reservoir computers, in order to predict high-dimensional spatiotemporally chaotic systems, as was first described in \cite{parallelreservoir2018} (but without the incorporation of a knowledge-based model). Finally, in Sec.~\ref{sec:hybrid-parallel}, we present our proposed CHyPP architecture and technique for combining the parallel reservoir method of Sec.~\ref{sec:parallelization} with the hybridization of a knowledge-based predictor and a parallel reservoir-based machine learning prediction of Sec.~\ref{sec:hybrid}. It is our belief that it is only by means of such a combination that the most effective application of machine-learning-enabled prediction can be realized for large, complex, spatiotemporally chaotic dynamical systems.

\subsection{Basic Reservoir Computing}\label{sec:reservoir-basic}
\begin{figure}[h!]
    \centering
    \includegraphics[width = 0.7\textwidth]{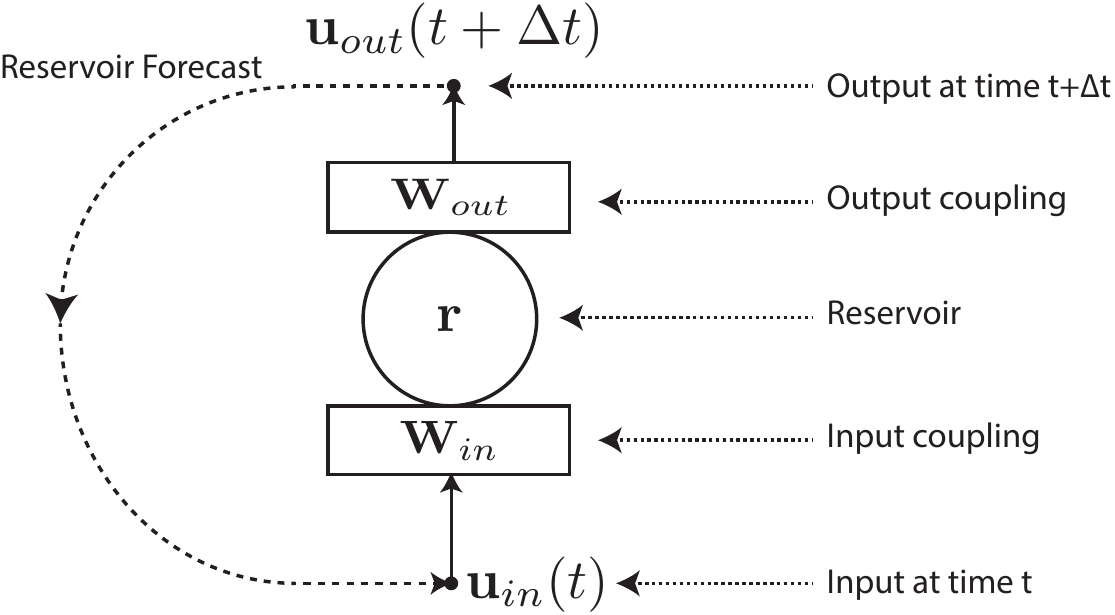}
    \caption{Diagram of the reservoir computer setup. In the ``open-loop'' training phase (analogous to Fig.~\ref{fig:basicML}(a)), the dashed line representing coupling from the output back to the input is absent. In the ``closed-loop'' prediction phase (analogous to Fig.~\ref{fig:basicML}(b)), the coupling from the output back to the input (dashed line) is activated.}
    \label{fig:reservoirdiagram}
\end{figure}

We now consider that the ML device shown in Fig.~\ref{fig:basicML} is a reservoir computer which, as shown in Fig.~\ref{fig:reservoirdiagram} (and further discussed subsequently), consists of a linear input coupler ($~\mathbf{W}_{in}$) that couples the state $\vv{u}_{in}(t)$ into the reservoir (the circle in Fig.~\ref{fig:reservoirdiagram}). The state of the reservoir is given by a high-dimensional vector $\vv{r}(t)$, which is then linearly coupled by $\vv{W}_{out}$ to produce an output vector $\vv{u}_{out}(t+\Delta t)$ which, through the training, is a very good approximation to $\vv{u}(t+\Delta t)$. In this paper, our implementation of the reservoir is an artificial recurrent neural network with a large number of nodes.
The artificial neural network that forms our basis of the reservoir computing implementation is illustrated in Fig.~\ref{fig:reservoirdiagram}. The reservoir network adjacency matrix is chosen to be a randomly generated, sparse matrix $\vv{A}$ that represents a directed graph with weighted edges. The adjacency matrix $\vv{A}$ has dimensions $D_r \times D_r$, where $D_r$ is the number of nodes in the network. Elements of $\vv{A}$ are randomly generated such that the average number of incident connections per node (average number of nonzero elements of the matrix in each row) is set to a chosen value $\langle d \rangle$, the ``average in-degree", while the nonzero elements of $\mathbf{A}$ are chosen from a uniform distribution over the interval $[-1,1]$. Once generated, $\vv{A}$ is re-scaled (i.e., multiplied by a scalar) so that the its largest absolute eigenvalue is a prescribed value $\rho$, called the spectral radius. Each node $i$ in the network has an associated scalar state $r_i(t)$. The state of the network is represented by the $D_r$ dimensional vector $\vv{r}(t)$, whose elements are $r_i(t)$ where $i=1,2,3,\dots ,D_r$.  

The reservoir network state $\vv{r}(t)$ evolves dynamically while receiving input through a $K \times D_r$ input coupling matrix, $\vv{W}_{in}$. We choose the matrix $\vv{W}_{in}$ to contain an equal number of nonzero elements in each column, which corresponds to coupling each element of the reservoir input to an equal number of reservoir node states. Nonzero elements of this matrix are selected randomly and uniformly from the interval [-$\sigma$, $\sigma$], where $\sigma$ is referred to as the input weight. Given the current state of the reservoir $\vv{r}(t)$, the reservoir state is advanced at each time using a hyperbolic tangent activation function,
\begin{align}\label{eq:reservoir_advance}
\vv{r}(t+\Delta t) = \tanh[\vv{Ar}(t) + \vv{W}_{in}\vv{u}_{in}(t)].
\end{align}

Before prediction begins, the reservoir computer is trained in the ``open-loop'' configuration. During this training phase, $\vv{u}_{in}(t) = \vv{u}(t)+s\boldsymbol{\eta}(t)$. Here, $\vv{u}(t)$ is the measurement of dynamical system state at time $t$ in the form of a $K$-dimensional vector. As in Ref.~\cite{herbert2001echo}, we add a small, normally distributed $K$-dimensional vector $s\boldsymbol{\eta}(t)$ of mean 0 and standard deviation $s$ to the input dynamical system state during training. The elements of the vector $\boldsymbol{\eta}(t)$ are chosen randomly and independently at each time $t$. The function of this added ``stabilization noise'' is to allow the reservoir computer to learn how to return to the true trajectory when the input trajectory has been perturbed away from it. We find that, in many cases, this additional small noise input beneficially promotes stability of the closed-loop prediction configuration once training has been completed.

The adjustable constants characterizing the overall prediction system (e.g., $D_r$, $\langle d \rangle$, $\sigma$, $s$, and $\Delta t$) are referred to as ``hyperparameters'', and it is important that they be chosen carefully in order for the reservoir computer to predict accurately. For example, as explained in previous literature, the hyperparameters can be chosen so that the reservoir system has the so-called ``echo-state property'' (see, e.g., Ref.~\cite{herbert2001echo}) whereby, when in the open-loop training phase, the reservoir state $\vv{r}(t)$, aside from an initial transient and with the random sequence $\boldsymbol{\eta}(t)$ fixed, the reservoir state $\vv{r}(t)$ becomes uniquely determined by the reservoir input sequence $\vv{u}(t)$ (and hence independent of the initial values of $\vv{r}$). Accordingly, prior to initiation of the training, we ignore and discard the reservoir and input states for the first few time steps. The state of the reservoir at the end of this transient nullification period is labelled $\vv{r}(0)$. Starting with $\vv{r}(0)$, the training system states $\vv{u}(j\Delta t)$ ($j$ an integer, $j\Delta t \leq t_0$) and the resulting reservoir states, $\vv{r}((j+1)\Delta t)$, are recorded and saved. We then desire to the use these saved states to produce an output, $\tilde{\vv{u}}(t+\Delta t)$, when $\vv{u}(t)$ is the input, which we desire to be very close to $\vv{u}(t+\Delta t)$. To do this, we find it useful to perform an ad-hoc operation on the reservoir state vectors that squares the value of half of the node states. Specifically, we define $\tilde{\vv{r}}(j\Delta t)$ such that,
\begin{align}\label{eq:rtilde}
    \tilde{r}_i &= r_i\textrm{ for }i\textrm{ odd,}\\
    \tilde{r}_i &= r_i^2\textrm{ for }i\textrm{ even.}
\end{align}
As surmised in footnote [16] of Ref.~\cite{parallelreservoir2018}, this operation improves prediction by breaking a particular odd symmetry of the reservoir dynamics that is not generally present in the dynamics to be predicted. We next couple the transformed reservoir state $\tilde{\vv{r}}(t+\Delta t)$ via a $K\times D_r$ output coupling matrix $\vv{W}_{out}$ to produce an output $\vv{u}_{out}(t+\Delta t)$,
\begin{align}\label{eq:output_matrix}
    \vv{u}_{out}(t+\Delta t) = \vv{W}_{out}\tilde{\vv{r}}(t+\Delta t), 
\end{align}
and we endeavor to choose (``train'') the matrix elements of $\vv{W}_{out}$ so that $\vv{u}_{out}(t+\Delta t)$ is close to $\vv{u}(t+\Delta t)$. In general, this will require that $D_r \gg K$. To accomplish this, we try to minimize the $L^2$ difference between $\vv{u}(t + \Delta t)$ and $\vv{u}_{out}(t+\Delta t)$. To prevent overfitting, we insert a Tikhonov regularization term to penalize very large values of the matrix elements of $\vv{W}_{out}$~\cite{tikhonov_solutions_1977}; that is, we find
\begin{align}\label{eq:res_cost}
    \underset{\vv{W}_{out} }{\min } \bigg\{\sum_{0\leq t < t_0}\Big[\lVert \vv{W}_{out}\tilde{\vv{r}}(t)-\vv{u}(t) \rVert^2 \Big] + \beta \mathrm{Trace}(\mathbf{W}_{out}\vv{W}_{out}^T)\bigg\},
\end{align}
over the $KD_r$ scalar values of the matrix $\vv{W}_{out}$. Here, $\beta$ is a small regularization parameter and $\lVert \dots \rVert$ denotes the $L^2$ norm. This technique is commonly known as ridge regression. In our subsequent numerical experiments (Secs.~\ref{sec:results} and~\ref{sec:multiscale}), we use the ``matrix solution'' for the minimization problem to determine the trained $\vv{W}_{out}$. In particular, we proceed as follows. We first form a matrix $\tilde{\vv{R}}$ where the $j^{th}$ column is the $j^{th}$ transformed reservoir state $\tilde{\vv{r}}(j\Delta t)$.
We define a target matrix $\vv{U}$ consisting of the time series of training data such that the $j^{th}$ column of $\vv{U}$ is $\vv{u}(j\Delta t)$.
We then determine a matrix $\vv{W}_{out}$ that satisfies the following linear system,

\begin{align}
    \label{eq:wout}
    \vv{W}_{out}(\tilde{\vv{R}} \tilde{\vv{R}}^{T} + \beta I) &= \vv{U} \Tilde{\vv{R}}^T.
\end{align}
We note that methods of solving Eq.(\ref{eq:res_cost}) for $\vv{W}_{out}$ other than direct matrix solution are also available and may sometimes be advantageous (e.g., GMRES~\cite{saad_gmres_1986}, stochastic gradient descent~\cite{zhang_solving_2004}, etc.). By means of this minimization, it is hoped that $\vv{u}_{out}(t) \cong \vv{u}(t)$ is achieved. This completes the training process, following which we can switch to the closed-loop configurations (Fig.~\ref{fig:basicML}b and the dashed line in Fig.~\ref{fig:reservoirdiagram}) and attempt to predict the subsequent evolution of $\vv{u}(t)$. Prediction can then proceed via Eqs.(\ref{eq:reservoir_advance}), (\ref{eq:rtilde}) and (\ref{eq:wout}) where the prediction of the dynamical system state $\tilde{\vv{u}}(t) = \vv{u}_{out}(t)$ and the reservoir input is received from the feedback loop ($\vv{u}_{in}(t) = \vv{u}_{out}(t)$).

The closed-loop configuration system can be regarded as a surrogate dynamical system that mimics the original unknown dynamical system. As such, if the  original unknown dynamical system is chaotic, the closed-loop predictor system will also be chaotic. Due to the exponential growth of small errors implied by chaos, we cannot expect prediction to be good for more than several Lyapunov times (the Lyapunov time is the typical e-folding time for error growth in a chaotic system). Thus we will regard our predictions to be successful when they are good for a few Lyapunov times.

Now consider that we have made a prediction for $\vv{u}(t)$, and, at some later time, we wish to perform another prediction of the same spatiotemporally evolving system with unknown dynamics. It is not necessary to retrain our predictor; we can, instead, re-use the previously obtained $\vv{W}_{out}$ ~\cite{parallelreservoir2018}. To do so, we re-initialize the reservoir state to zero, switch the reservoir computer into its open-loop configuration, and allow it to evolve given input states of the unknown dynamical system measured at times $t_p-T_S \leq t \leq t_p$ (i.e., $\vv{u}_{in}(t) = \vv{u}(t)$). $T_S$ is some synchronization time that is sufficiently longer than the characteristic memory of the reservoir computer but, importantly, is much shorter than the necessary training time needed to determine $\vv{W}_{out}$. $t_p$ is the time at which we want to begin our prediction. After this synchronization period, the reservoir computer is switched to its standard closed-loop prediction configuration and is used to make predictions at later times.

If the original system is very high dimensional (i.e., the dimension $K$ of the measure vector $\vv{u}(t)$ is very large), then $D_r \gg K$ must be exceedingly large. This can make the training to determine $\vv{W}_{out}$ infeasible. For example, if we solve Eq.(\ref{eq:res_cost}) by the direct matrix method, Eq.(\ref{eq:wout}) shows that we must solve a $D_r\times D_r$ linear system of equations. For our computational resources, we find that this becomes impossible as $D_r$ approaches $~2\times10^4$. Due to this and other similar considerations, we deem the method discussed in this section to be untenable for the prediction of large, spatiotemporally chaotic systems of the type we are interested in.
\subsection{Hybrid Reservoir Computing}
\label{sec:hybrid}
\begin{figure}[h]
    \centering
    \includegraphics[width = 0.7\textwidth]{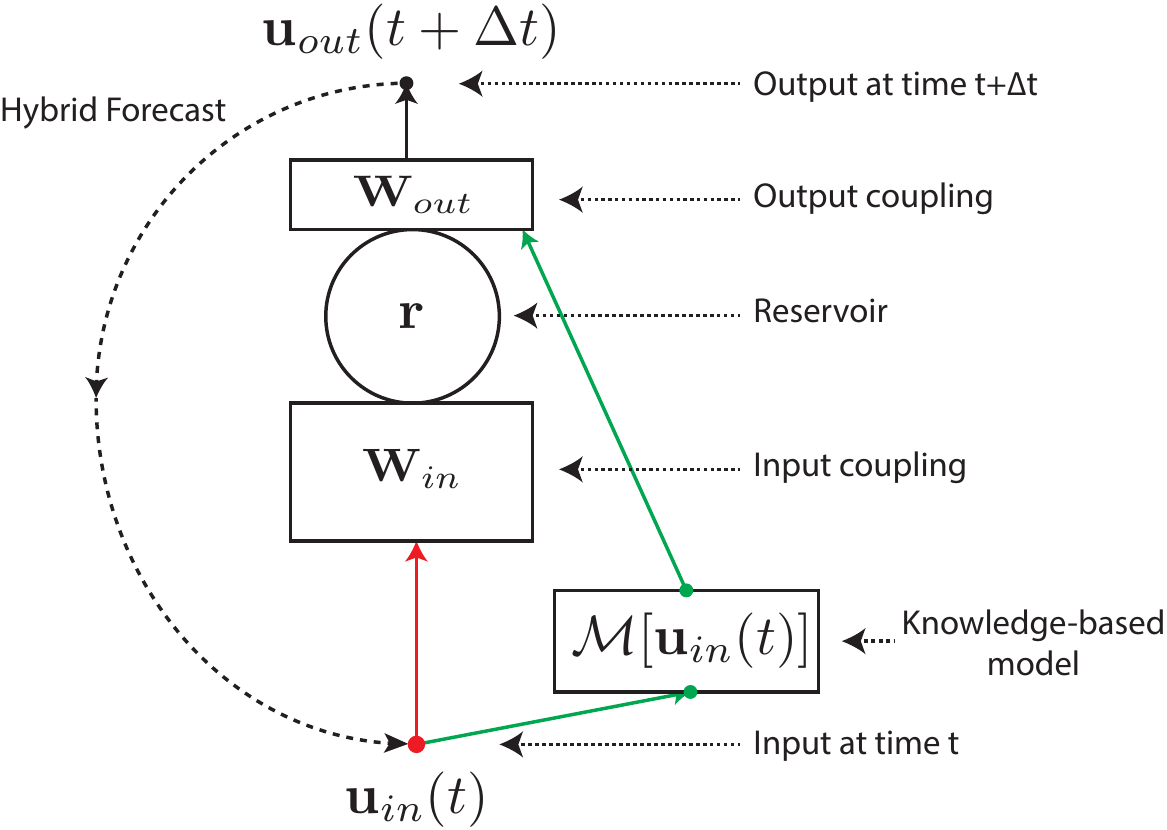}
    \caption{Diagram of the hybrid reservoir computer setup. In the ``open-loop'' training phase (analogous to Fig.~\ref{fig:basicML}(a)), the dashed line representing coupling from the output back to the input is absent. In the ``closed-loop'' prediction phase (analogous to Fig.~\ref{fig:basicML}(b)), the coupling from the output back to the input (dashed line) is activated.}
    \label{fig:hybriddiagram}
\end{figure}
In this section, we briefly review a hybrid scheme proposed in Ref.~\cite{hybridreservoir2018} for combining reservoir computing with an imperfect knowledge-based model of the dynamical system of interest. We again assume access to time series data of measurements of the state of the dynamical system. We further assume that an imperfect knowledge-based model of the system producing the measurements is available and that this imperfect model is capable of forecasting the state of the dynamical system with some degree of accuracy, which we wish to improve upon. In the hybrid setup of Ref.~\cite{hybridreservoir2018} (Fig.~\ref{fig:hybriddiagram}) described below, it has been shown that the machine learning method and the knowledge-based model augment each other and, in conjunction, can provide a significantly better forecasts than either the knowledge-based model or the pure machine learning model acting alone. 

As in Sec.~\ref{sec:reservoir}, we assume that the data used for training is given by $K$ measurements of the state of the dynamical system at equally spaced increments in time, $\Delta t$, forming a vector time series $\vv{u}(t)$. The imperfect knowledge-based model $\mathcal{M}$ is an operator that maps the state $\vv{u}(t)$ to a forecast of the state at time $(t + \Delta t)$.

We advance the reservoir state in time using the same activation function as described in Sec.~\ref{sec:reservoir-basic},
\begin{align}\label{eq:reservoir_advance_hybrid}
    \vv{r}(t + \Delta t) &= \tanh[ \vv{A}\vv{r}(t) + \vv{W}_{in}\vv{u}_{in}(t)].
\end{align}
Once again, during the training phase, $\vv{u}_{in}(t) = \vv{u}(t)+s\boldsymbol{\eta}(t)$. The training process is similar to the one employed in Sec.~\ref{sec:reservoir} for the basic reservoir computer but with the addition of the knowledge-based prediction (as illustrated in Fig.~\ref{fig:hybriddiagram}). Using ridge regression, we find a linear mapping $\vv{W}_{out}$ from $\tilde{\vv{r}}(t)$ and $\mathcal{M}[\vv{u}(t)+s\boldsymbol{\eta}(t)]$ to produce an approximate prediction of $\vv{u}(t+\Delta t)$,

\begin{align}
    \label{eq:hybridopt}
    \vv{W}_{out}\begin{bmatrix}
        \tilde{\vv{r}}(t+\Delta t)\\
        \mathcal{M}[\vv{u}_{in}(t)]
        \end{bmatrix}
        \simeq \vv{u}(t + \Delta t).
\end{align}
Here, $\vv{u}_{in}(t)$ is the same as that input to the reservoir, $\vv{u}_{in}(t) = (\vv{u}(t)+s\boldsymbol{\eta}(t)$. We again include the small $s\boldsymbol{\eta}(t)$ vector in the knowledge-based model input during training to improve the stability of the method. Additionally, recall that $\tilde{\vv{r}}$ is related to $\vv{r}$ by Eq.(\ref{eq:rtilde}). In the prediction phase, we run the hybrid system in a closed loop feedback configuration (Fig.~\ref{fig:hybriddiagram} with the dashed line feedback connection present) using Eqs.(\ref{eq:reservoir_advance_hybrid}), (\ref{eq:rtilde}), and the following equation,

\begin{align}
    \vv{u}_{out}(t+\Delta t) &=  \vv{W}_{out}\begin{bmatrix}
    \Tilde{\vv{r}}(t+\Delta t)\\
    \mathcal{M}[\vv{u}_{out}(t)]\end{bmatrix}.
\end{align}
During the prediction phase, the hybrid forecast $\tilde{\vv{u}}_\mathcal{H}(t) = \vv{u}_{out}(t)$ and the hybrid input is received from the feedback loop ($\vv{u}_{in}(t) = \vv{u}_{out}(t)$). Note that, in this scheme, the output is a linear combination of the reservoir state and the knowledge-based model output that optimizes the agreement of the combined system output with the training data. Thus, we can regard the result as being an optimum combination of the reservoir and knowledge-based components. Hence, we expect that if one component is superior for some aspect of the prediction, then it will be weighted more highly for that aspect of the prediction. This suggests that predictions by this method may be greatly improved over those available from either the knowledge-based component or the reservoir component acting alone (e.g., see Fig. 7 of Ref.~\cite{hybridreservoir2018}).

In addition to the hybrid configuration shown in Fig.~\ref{fig:hybriddiagram}, we have also tested a modified configuration in which there is an additional input to the 
reservoir component from the output of the knowledge-based model $\mathcal{M}$. We have empirically found that this modification sometimes yields a small positive improvement in prediction; however, for simplicity, we henceforth only consider the configuration in the figure.

\subsection{Parallelization}
\label{sec:parallelization}
\begin{figure}[h!]
    \centering
    \includegraphics[width = 0.9\textwidth]{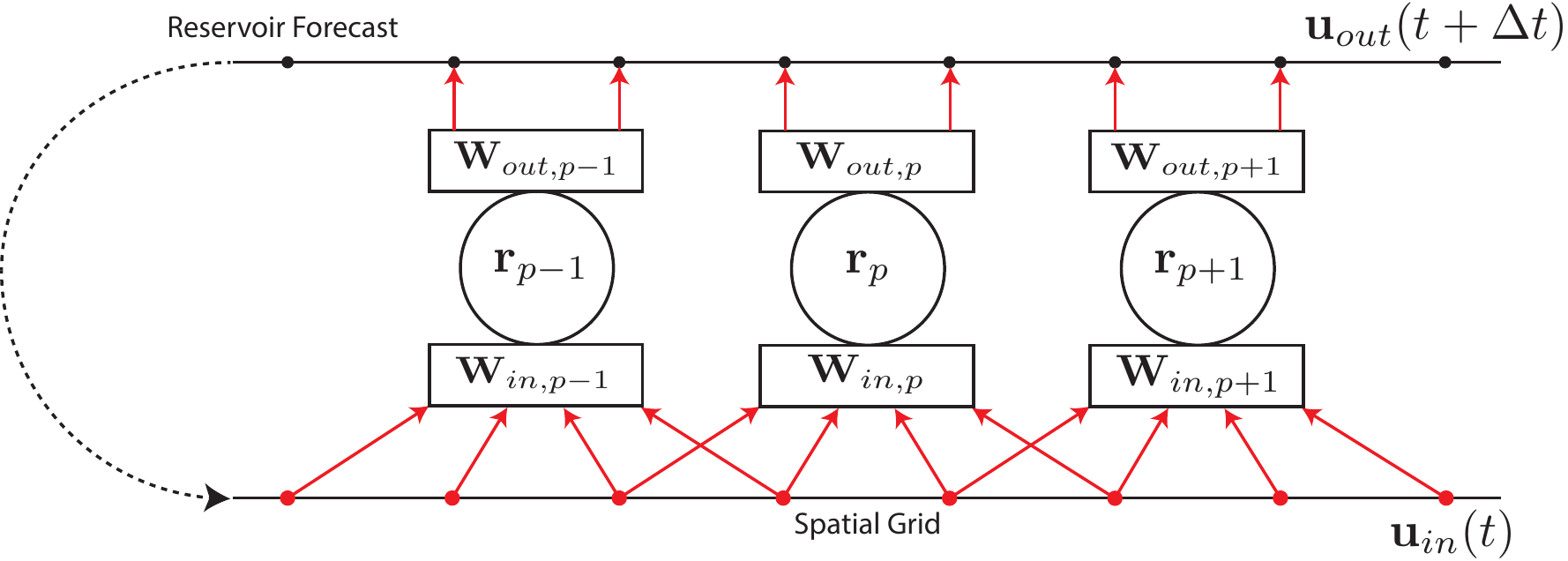}
    \caption{Diagram of parallel reservoir computer setup.}
    \label{fig:paralleldiagram}
\end{figure}
To obtain a good prediction of a chaotic dynamical system state using reservoir computing, the reservoir dimensionality must be much greater than that of the dynamical system (i.e., $D_r \gg K$) so that there are enough free parameters available in $\vv{W}_{out}$ for fitting the reservoir output state to the measured dynamical system state at time $(t+\Delta t)$. This can cause the computational cost of determining an optimum output matrix to become unfeasibly high for large dynamical systems, e.g. because implementation of this step by the method of Eq.(\ref{eq:wout}) involves solving a $D_r \times D_r$ linear system. As a point of reference, we note that the dimension of the state vector of a current typical operational global weather forecasting models is on the order of $10^9$. A method to make consideration of such problems feasible for machine learning approaches was proposed in Ref.~\cite{parallelreservoir2018}. The idea is to exploit the short range of causal interactions over a small time period in  many spatiotemporally chaotic dynamical systems. This was shown to allow the use of multiple relatively small reservoir computers that each make predictions of a part of the full dynamical system state as in a local region, illustrated in Fig.~\ref{fig:paralleldiagram} and explained below. This method has the advantage that, in the training phase, all of the relatively small output matrices $\vv{W}_{out,p}$ of each reservoir computer can be trained independently in parallel, thus greatly reducing the difficulty of training.

For illustrative purposes, consider a spatiotemporal dynamical system in one spatial dimension with periodic boundary conditions. Let the dynamical system state be represented by a $K$-dimensional vector time series  $\vv{u}(t) = [u_1(t), u_2(t), \dots, u_K(t)]^T$ where each scalar component $u_j(t)$ represents the time series at a single spatial grid point. We divide the system state into $P$ equally sized, contiguous regions containing $Q$ system variables, where $PQ = K$. We denote the system variables in these regions as $\vv{u}_p(t) = [u_{Q(p-1)+1}(t), \dots, u_{Qp}(t)]^t$ where $1\leq p\leq P$. Each local region in space is predicted by a reservoir $R_p$, each of which has internal reservoir states $\vv{r}_p(t)$ and adjacency matrix $\vv{A}_p$ generated via the process described in Sec.~\ref{sec:reservoir-basic}. Each reservoir is coupled to its input, $\vv{u}_{in,p}(t)$, via a matrix of input weights, $\vv{W}_{in,p}$. This input corresponds to a local region of the system that contains the region to be predicted by that reservoir as well as a size $\ell$ overlap region on either side, $\vv{u}_{in,p}(t) = [u_{in,Q(p-1)+1-\ell}(t),u_{in,Q(p-1)+2-\ell}(t),\dots, u_{in,Qp+\ell}(t)]^T$. We denote the dynamical system state in this input region by the size $(2\ell+Q)$ dimensional vector $\vv{v}_p(t) = [u_{Q(p-1)+1-\ell}(t),u_{Q(p-1)+2-\ell}(t),\dots, u_{Qp+\ell}(t)]^T$. This overlap accounts for the short range causal interactions across the boundaries of the local regions. The assumption here is that, over the incremental prediction time $\Delta t$, state information does not propagate fast enough for nodes outside the input regions of reservoir $p$ to influence the time $\Delta t$ change in the dynamical system states predicted by reservoir $p$. 

Each reservoir state is advanced using the following equation,
\begin{align}\label{eq:reservoir_advance_parallel}
\vv{r}_p(t+\Delta t) = \tanh[\vv{A}_p\vv{r}_p(t) + \vv{W}_{in,p}\vv{u}_{in,p}(t)].
\end{align}
During the training phase (Fig.~\ref{fig:paralleldiagram} with the dashed output-to-input connection absent), $\vv{u}_{in,p}(t) = \vv{v}_p(t)+s\boldsymbol{\eta}(t)_p$
Here, the $(2\ell + Q)$ dimensional vector $\boldsymbol{\eta}(t)_p$ is the $p^{th}$ local region of a global vector of normally distributed random variables, $\boldsymbol{\eta}(t)$, chosen independently at each time $t$, 
\begin{align}\label{eq:localeta}
\boldsymbol{\eta}(t)_p = \begin{bmatrix}\eta_{Q(p-1)+1-\ell}\\\eta_{Q(p-1)+2-\ell}\\\vdots\\\eta_{Qp+\ell}\end{bmatrix}.
\end{align}
After a suitably long transient nullification period, we determine the output matrices $\vv{W}_{out,p}$ for each reservoir that solve the least squared optimization problem using ridge regression,
\begin{align}\label{eq:parallelres_cost}
    \underset{\vv{W}_{out,p} }{\min } \bigg\{\sum_{0\leq t < t_0}\Big[\lVert \vv{W}_{out,p}\tilde{\vv{r}_p}(t+\Delta t)-\vv{u}_p(t+\Delta t) \rVert^2 \Big] + \beta \mathrm{Trace}(\mathbf{W}_{out,p}\vv{W}_{out,p}^T)\bigg\}.
\end{align}
Note that, for each $p$, the matrix $\vv{W}_{out,p}$ can be relatively small as the number of outputs is the number of state variables in region $p$ (not the entire global state). Furthermore, the determinations of the relatively small $\vv{W}_{out,p}$ matrices are independent for each region $p$, and thus can be computed in parallel. The ``direct matrix method'' solution for determining each of the $\vv{W}_{out,p}$ matrices proceeds as follows. First, we rewrite Eq.(\ref{eq:parallelres_cost}) as 
\begin{align}
    \label{eq:parallelmin}
    \underset{\vv{W}_{out,p} }{\min } \Vert \vv{U}_p-\vv{W}_{out,p}\Tilde{\vv{R}}_p \rVert^2 + \beta \mathrm{Trace}(\vv{W}_{out,p}\vv{W}_{out,p}^T),
\end{align}
where, in Eq.(\ref{eq:parallelmin}), $\vv{U}_p$ and $\Tilde{\vv{R}}_p$ are analogous to $\vv{U}$ and $\Tilde{\vv{R}}$ in the single reservoir prediction (see Eq.(\ref{eq:wout})). $\vv{U}_p$ is the target matrix such that the $j^{th}$ column is $\vv{u}_p(j\Delta t)$, while $\Tilde{\vv{R}}_p$ is obtained from $\vv{R}_p$ analogous to the single reservoir case as described in Eq.(\ref{eq:rtilde}). Each $\vv{W}_{out,p}$ is calculated by solving the following equation,
\begin{align}
    \label{eq:woutparallel}
    \vv{W}_{out,p}(\tilde{\vv{R}}_{p} \tilde{\vv{R}}_{p}^{T} + \beta I) &= \vv{U}_{p} \Tilde{\vv{R}}^T_{p}.
\end{align}
Note that, as previously claimed, the minimization problem for each $p$, Eq.(\ref{eq:parallelmin}), is completely independent, and can be solved for the different $p$ in parallel. As in Sec.~\ref{sec:reservoir-basic}, in the prediction phase and, after a period of synchronization, we produce a full state prediction $\tilde{\vv{u}}(t)$ by running the system in a closed loop feedback configuration (i.e., Fig.~\ref{fig:paralleldiagram} with the dashed output-to-input feedback connection present). This is done by concatenating the local predictions from each reservoir $\tilde{\vv{u}}_p(t) = \vv{u}_{out,p}(t)$ (where $\vv{u}_{out,p}(t)$ is the output state from each reservoir reservoir). Reservoir $p$ then receives inputs from its own outputs in addition to the left and right overlap zone inputs from the $\ell$ left grid points, and the $\ell$ right grid points. The entire system thus evolves as follows,
\begin{align}
    \vv{u}_{out,p}(t) &= \vv{W}_{out,p}[\Tilde{\vv{r}}_p(t)],\\
    \Tilde{\vv{u}}(t) &= \begin{bmatrix}\Tilde{\vv{u}}_1(t)\\ \Tilde{\vv{u}}_2(t)\\\vdots\\ \Tilde{\vv{u}}_P(t)\end{bmatrix}=\begin{bmatrix}\vv{u}_{out,1}(t)\\ \vv{u}_{out,2}(t)\\\vdots\\ \vv{u}_{out,P}(t)\end{bmatrix},\quad
    \vv{u}_{in,p}(t) = \Tilde{\vv{v}}_p(t) =\begin{bmatrix} \Tilde{u}_{Q(p-1)+1-\ell}(t)\\\Tilde{u}_{Q(p-1)+2-\ell}(t)\\\vdots\\ \Tilde{u}_{Qp+\ell}(t)\end{bmatrix},\\
    \vv{r}_p(t+\Delta t) &= \tanh[\vv{A}_p\vv{r}_p(t)+\vv{W}_{in,p}\vv{u}_{in,p}(t)].
\end{align}
$\tilde{\vv{r}}_p(t)$ is obtained from $\vv{r}_p(t)$ using Eq.(\ref{eq:rtilde}).
\subsection{Combined Hybrid/Parallel Prediction (CHyPP)}
\label{sec:hybrid-parallel}
\begin{figure}[h]
    \centering
    \includegraphics[width = 0.9\textwidth]{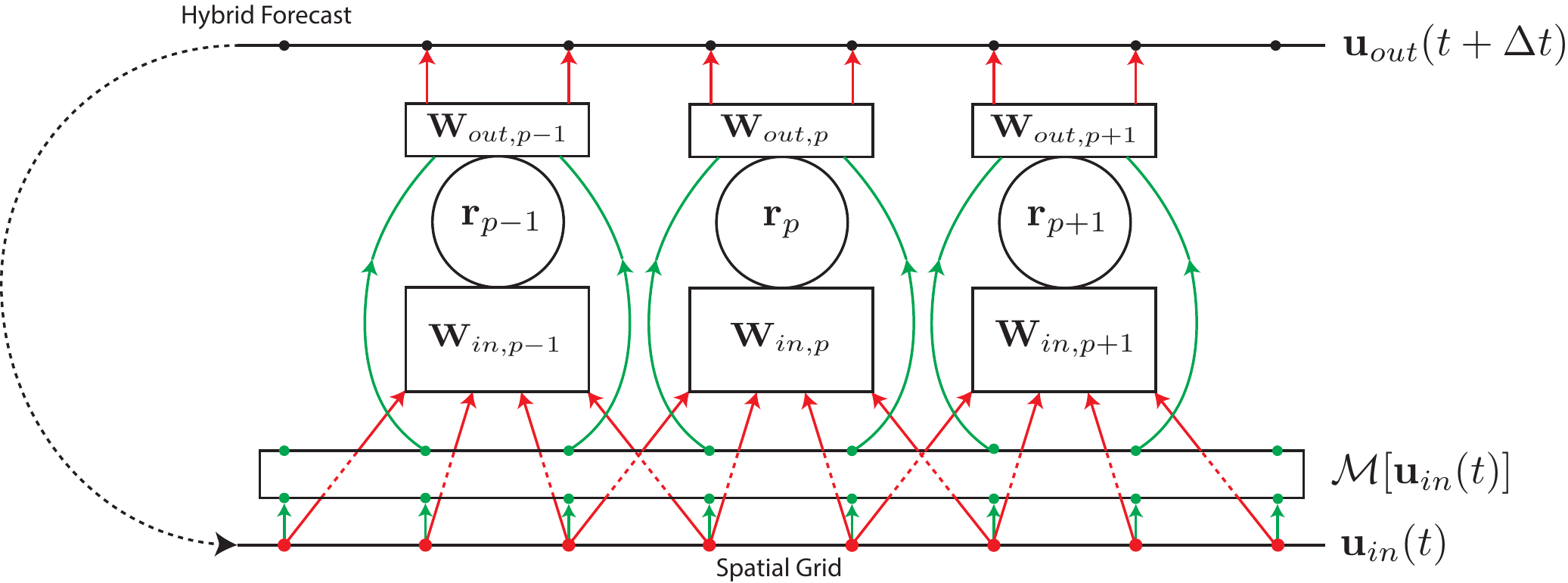}
    \caption{Diagram of the Combined Hybrid/Parallel Prediction (CHyPP) architecture using reservoir computing.}
    \label{fig:parallel_hybrid}
\end{figure}

In our previous work~\cite{hybridreservoir2018}, we constructed a hybrid prediction method using a single reservoir. In this section, we consider a combination of the parallel approach of Sec.~\ref{sec:parallelization} (which enables computationally efficient scaling to high-dimensional spatiotemporal chaos) and the hybrid approach of Sec.~\ref{sec:hybrid} (which allows us to utilize partial prior knowledge of the dynamical system) where we assume that the knowledge-based system provides \textit{global} predictions over the entire spatial domain. While the approach is easily generalized to 2- and 3-dimensional spatial processes, in order to most simply demonstrate our proposed methodology, we again consider a one-dimensional, spatiotemporally chaotic dynamical system with periodic boundary conditions, with a state represented by a $K$-dimensional vector time series $\vv{u}(t) = [u_1(t), u_2(t), \dots, u_K(t)]^T$. Our approximate knowledge-based prediction operator $\mathcal{M}$ gives a global prediction of the full system state for a time $\Delta t$: $\mathcal{M}[\vv{u}(t)] = \hat{\vv{u}}(t+\Delta t)$. As in our parallel reservoir computer prediction described in Sec.~\ref{sec:parallelization}, we partition the system state into $P$ equally sized, continuous regions containing $Q$ variables, where $PQ = K$ and each such region is predicted by a reservoir $R_p$, $p=1,2,\dots,P$.

Each reservoir $R_p$ input is coupled to a local region of the system states as in Sec.~\ref{sec:parallelization}, and the reservoir state $\vv{r}_p(t)$ is advanced using the following equation,
\begin{align}\label{eq:reservoir_advance_hybridparallel}
\vv{r}_p(t+\Delta t) = \tanh[\vv{A}_p\vv{r}_p(t) + \vv{W}_{in,p}\vv{u}_{in,p}(t)].
\end{align}
During the initial training phase, $\vv{u}_{in,p}(t) = \vv{v}_p(t)+s\boldsymbol{\eta}(t)_p$. In Eq.(\ref{eq:reservoir_advance_hybridparallel}), $\vv{W}_{in,p}$ is the input coupling matrix for the local system states, analogous to that described in Sec.~\ref{sec:reservoir-basic}. As in Sec.~\ref{sec:parallelization}, $\vv{v}_p(t)$ is the state measurements at grid points within the local region to be predicted along with $\ell$ grid points to either side and $\boldsymbol{\eta}(t)_p$ is the $p^{th}$ local region of the global vector of normally distributed random numbers $\boldsymbol{\eta}(t)$. Each reservoir is trained independently in parallel using a set of training data consisting of an equally spaced time series of measured states of the large scale dynamics beginning at $t=0$ after some initial transient nullification period. Again, we solve the least squares optimization problem with ridge regression to determine an output mapping for each reservoir (analogous to Eq.(\ref{eq:parallelmin})),
\begin{align}
    \label{eq:hybridparallelmin}
    \underset{\vv{W}_{out,p} }{\min } \bigg\{ \lVert \vv{U}_p-\vv{W}_{out,p}\begin{bmatrix}\Tilde{\vv{R}}_p\\\hat{\vv{U}}_p \end{bmatrix}\rVert^2 + \beta \mathrm{Trace}(
    \vv{W}_{out,p} \vv{W}_{out,p}^T)\bigg\}.
\end{align}
In Eq.(\ref{eq:hybridparallelmin}), $\hat{\vv{U}}_p$ is a matrix whose $j^{th}$ column is $\hat{\vv{u}}_p(j\Delta t)$, where $\hat{\vv{u}}_p(j\Delta t)$ is the knowledge-based prediction of the $p^{th}$ local region of the system, 
\begin{align}
    \label{eq:localmodel}
    \hat{\vv{u}}_p(t+\Delta t) = \begin{bmatrix}\mathcal{M}[\vv{u}(t)+s\boldsymbol{\eta}(t)]_{Q(p-1)+1}\\ \mathcal{M}[\vv{u}(t)+s\boldsymbol{\eta}(t)]_{Q(p-1)+2}\\
    \vdots\\
    \mathcal{M}[\vv{u}(t)+s\boldsymbol{\eta}(t)]_{Qp}\end{bmatrix}.
\end{align}
The solution to Eq.(\ref{eq:hybridparallelmin}) by the direct matrix method is

\begin{align}
    \label{eq:woutparallelhybrid}
    \vv{W}_{out,p}\big(\begin{bmatrix}\Tilde{\vv{R}}_p\\\hat{\vv{U}}_p \end{bmatrix} [\tilde{\vv{R}}^T_{p},\hat{\vv{U}}^T_p] + \beta I\big) &= \vv{U}_{p} [\tilde{\vv{R}}^T_{p},\hat{\vv{U}}^T_p].
\end{align}

In the prediction phase, we run the system in a closed-loop configuration according to Eqs.(\ref{eq:hybridparallelpred}) to predict each local region of the system given each reservoir state and knowledge-based model prediction, obtaining $\Tilde{\vv{u}}_p(j\Delta t)$. The local predictions are appropriately concatenated to form a full state prediction, which is then used as input for the next prediction step, as follows,

\begin{align}\label{eq:hybridparallelpred}
    \Tilde{\vv{u}}_p(t) = \vv{u}_{out,p}(t)&= \vv{W}_{out,p}\begin{bmatrix}\Tilde{\vv{r}}_p(t)\\ \hat{\vv{u}}_p(t)]\end{bmatrix},\\
    \Tilde{\vv{u}}(t) = \begin{bmatrix}\Tilde{\vv{u}}_1(t)\\ \Tilde{\vv{u}}_2(t)\\\vdots\\ \Tilde{\vv{u}}_P(t)\end{bmatrix},&\quad
    \vv{u}_{in,p}(t) = \Tilde{\vv{v}}_p(t) = \begin{bmatrix}\Tilde{u}_{Q(p-1)+1-\ell}(t)\\\Tilde{u}_{Q(p-1)+2-\ell}(t)\\
    \vdots\\ \Tilde{u}_{Qp+\ell}(t)]\end{bmatrix}\\
    \hat{\vv{u}}_p(t+\Delta t) &= \begin{bmatrix}\mathcal{M}[\vv{u}_{in}(t)]_{Q(p-1)+1}\\\mathcal{M}[\vv{u}_{in}(t)]_{Q(p-1)+2}\\\vdots\\\mathcal{M}[\vv{u}_{in}(t)]_{Qp}\end{bmatrix},\\
    \vv{r}_p(t+\Delta t) &= \tanh[\vv{A}_p\vv{r}_p(t) +  \vv{W}_{in,p}\vv{u}_{in,p}(t)].
\end{align}
In the above equations, we once again calculate $\tilde{\vv{r}}_{p}(t)$ from $\vv{r}_{p}(t)$ using Eq.(\ref{eq:rtilde}).
\section{Test Results on the Kuramoto-Sivashinsky Equation}\label{sec:results}

We test the effectiveness of our proposed CHyPP method (Sec.~\ref{sec:hybrid-parallel}) by forecasting the state evolution of the Kuramoto-Sivashinsky equation with periodic boundary conditions ~\cite{kuramoto1978,sivashinsky1977},
\begin{align}\label{eq:KS}
    \frac{\partial y}{\partial t} = -y\frac{\partial y}{\partial x} - \frac{\partial^2 y}{\partial x^2}-\frac{\partial^4y}{\partial x^4},
\end{align}
where $y=y(x,t)$ and $y(x+L,t)=y(x,t)$. This spatially one-dimensional model generally produces spatiotemporally chaotic dynamics for periodicity length $L \gtrsim 50$. For the purpose of comparing various methods, we regard Eq.(\ref{eq:KS}) as generating the state measurements of a putative system that we are interested in, while the imperfect prediction model we use is a modified version of this same equation, where an error term $\epsilon$ is introduced in the coefficient of the second derivative term,
\begin{align}\label{eq:KS_werror}
     \frac{\partial y}{\partial t} = -y\frac{\partial y}{\partial x} - (1+\epsilon)\frac{\partial^2 y}{\partial x^2}-\frac{\partial^4y}{\partial x^4},
\end{align}
We have also investigated the case where the error is introduced by multiplying the $y\partial y/\partial x$ term in Eq.(\ref{eq:KS}) by $(1+\epsilon)$. For the latter case, the results of our method are qualitatively similar to those for Eq.(\ref{eq:KS_werror}). We form our simulated measured time series $\vv{u}(t)$ by taking the $i^{th}$ element of $\vv{u}(t)$ to be $y(i\Delta x,t)$, where $\Delta x = L/K$ is the grid spacing used for our numerical solutions of Eq.(\ref{eq:KS}). As a metric for how long a prediction is valid, we calculate the normalized root-mean-square error (NRMSE) between the true and predicted system states. We define the length of valid prediction, or ``valid time'', to be the time at which the NRMSE exceeds 0.2. Since the NRMSE saturates at $sqrt(2)$, we consider this to be the point when error in the prediction reaches about 15\% of its saturation value. In Sec.~\ref{sec:scalability}, we demonstrate that our CHyPP methodology can scale to predict very large systems. In Sec.~\ref{sec:traininglength}, we show that the CHyPP method requires significantly less training data than the parallel scheme of Sec.~\ref{sec:parallelization} (using reservoirs without a knowledge-based component).  We then discuss, in Sec.~\ref{sec:locality}, the sensitivity of the CHyPP and parallel machine learning (Sec.~\ref{sec:parallelization}) methods to the local overlap length $\ell$.

For all of our numerical experiments in this paper, in addition to our solution of the imperfect model, we use a digital computer to implement the machine learning. However, if, in the future, CHyPP is applied to very large systems (e.g., weather forecasting) requiring a much larger number of parallel machine learning units, then we envision that it may prove useful to perform the parallel machine learning using a special purpose physically implemented reservoir computing array, e.g., based on FPGA's or photonic devices~\cite{penkovsky_efficient_2018,tanaka_recent_2019,vinckier_high-performance_2015}. Such implementations, called ``AI hardware accelerators'', show great promise with respect to low cost, speed, and compactness.

\subsection{Prediction Scalability}\label{sec:scalability}
\begin{table}[h!]
\centering
\begin{tabular}{|c|c||c|c|}
    \hline
    $\langle d \rangle$ (average in-degree) & 3 & & \\
    \hline
     $\rho$ (spectral radius) & 0.6 & $\ell$ (local overlap length) & 6 \\
     \hline
     $\sigma$ (input coupling strength) & 0.1  & $\Delta t$ (Prediction time step) & 0.25\\ \hline
     $\beta$ (regularization) & $10^{-6}$ & $T_S$ (synchronization time) & 25\\ \hline
     $s$ (reservoir(s)-only tests) & $0.001$ & $s$ (CHyPP tests)& $0$\\
     \hline
\end{tabular}
\caption{Hyperparameters}
\label{tab:params1}
\end{table}
\begin{figure}[h!]
    \centering
    \includegraphics[width = \textwidth]{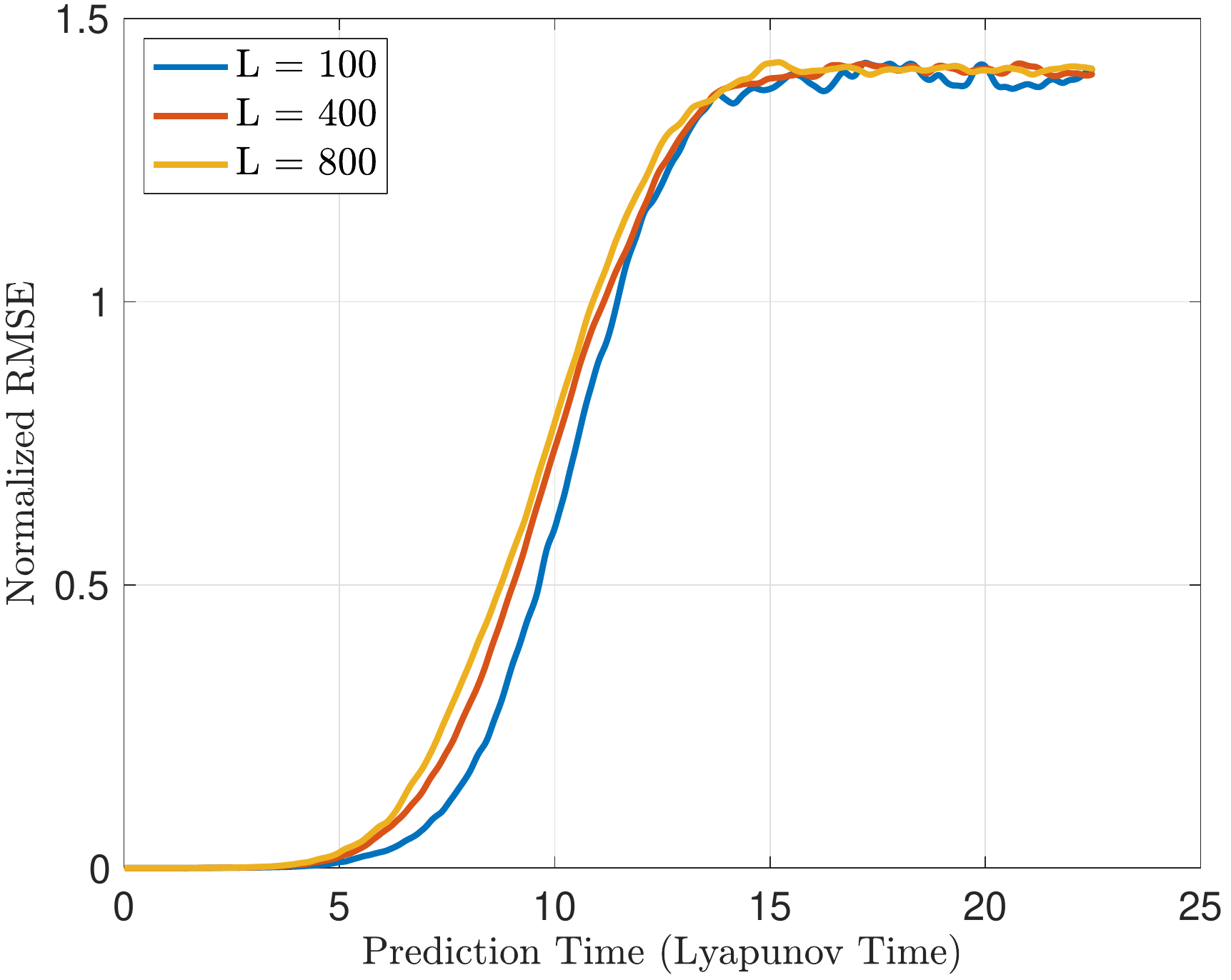}
    \caption{Average normalized root-mean-square error (NRMSE) in the CHyPP prediction where we have fixed the number of system variables predicted in each region to $Q=8$ and the reservoir spatial density at $P/L = 16/100$. Predictions are made using an imperfect model with second derivative error $\epsilon = 0.1$. The results shown are the average of 100 predictions where each prediction is made using the same set of reservoirs and training data set but where the CHyPP method is synchronized to different initial conditions for $\vv{u}(t)$. we see that the NRMSE curves are relatively invariant as $L$ increases, indicating that our method can be scaled to large systems.}
    \label{fig:rmsescalability}
\end{figure}
We first test the ability of our CHyPP method to scale to large system sizes. We consider the Kuramoto-Sivashinsky equation where we fix the number of system variables (grid points) each reservoir is trained to predict to $Q=8$, as well as the reservoir spatial density to $P/L = 16/100$ (P is the number of reservoirs), while varying the periodicity length $L$. For all tests in this section, we use the hyperparameters in Table~\ref{tab:params1} unless otherwise specified. We additionally fix the error in the incorrect model to $\epsilon = 0.1$. Fig.~\ref{fig:rmsescalability} shows the resulting NRMSE between the true state and our hybrid parallel prediction averaged over 100 prediction periods versus time. For each value of $L$ plotted in Fig.~\ref{fig:rmsescalability}, the density of the parallel reservoir computers is kept constant at $P/L = 0.16$. Additionally, the time plotted horizontally is in units of the Lyapunov time (the average chaos-induced e-folding time of small errors in the predicted state orbit). The NRMSE is relatively unchanged as the value of $L$ is increased, indicating that the CHyPP prediction, like the parallel reservoir-only prediction in~\cite{parallelreservoir2018}, can be scaled to very large systems by the addition of more reservoirs. 

Finally, we note that our test system, the Kuramoto-Sivashinky equation, is homogeneous in space, while the real systems we are interested in are generally spatially inhomogeneous. For example, weather forecasting accounts for geographic features (continents, mountains, etc.), as well as for latitudinal variation of solar input, among other spatially inhomogeneous factors. In order to ensure that our numerical tests with a homogeneous model are also relevant for a typical inhomogeneous situation, we have not made any use of the homogenaity of Eq.(\ref{eq:KS}): the adjacency matrices $\vv{A}_p$ corresponding to each reservoir $p$ are independently randomly generated, and each $\vv{W}_{out,p}$ is determined separately (rather than taking all $\vv{A}_p$ and $\vv{W}_{out,p}$ to be the same, as would be possible for a homogeneous system).
\subsection{CHyPP Promotes the Possibility of Good Performance Using a Relatively Small Duration of Training Data}\label{sec:traininglength}
\begin{figure}
    \centering
    \includegraphics[width = \textwidth]{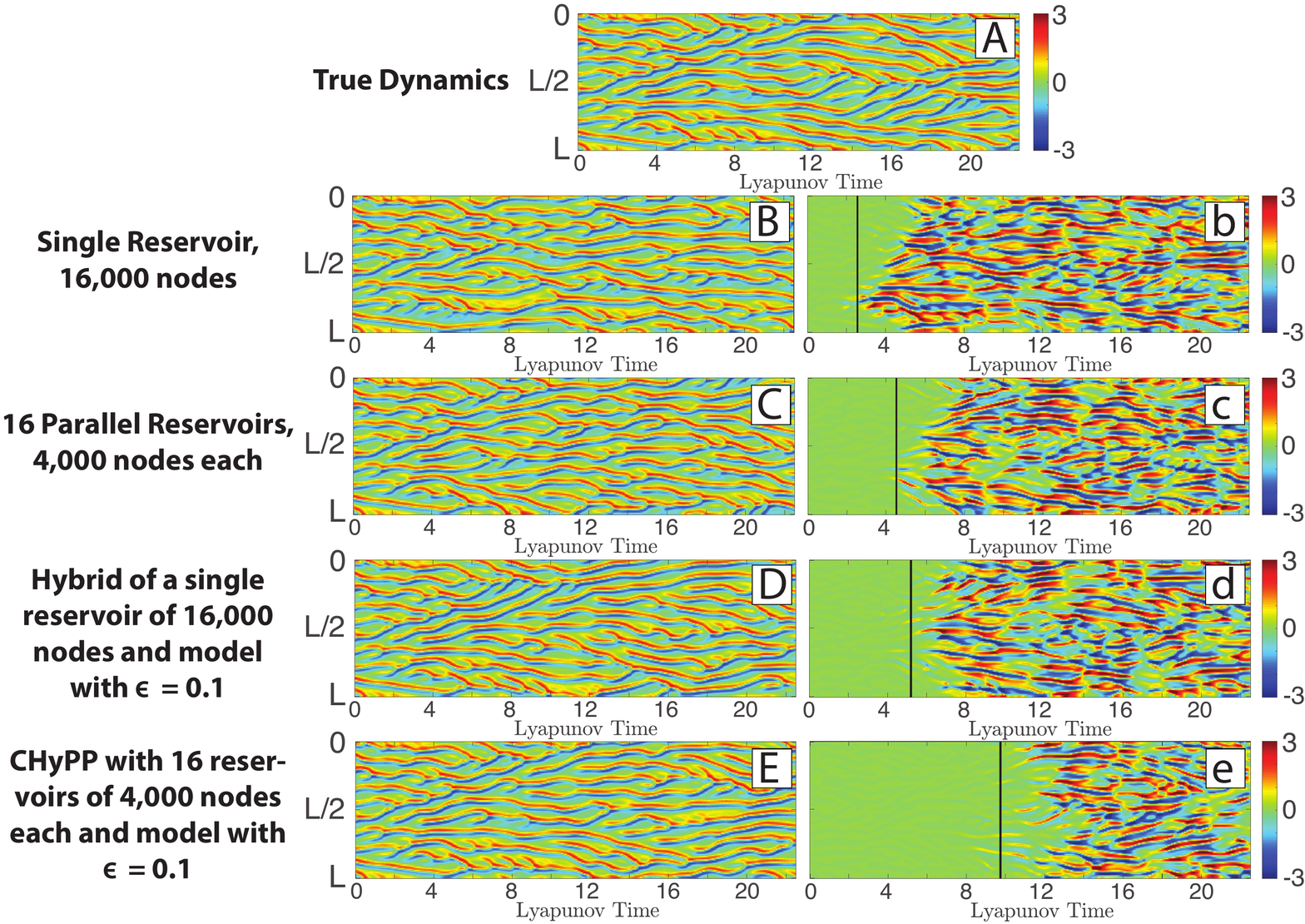}
    \caption{(A) The true dynamics of the Kuramoto-Sivashinsky equation obtained by numerical evolution of Eq.(\ref{eq:KS}) normalized to have mean 0 and variance 1. We plot the spatial grid point along the vertical axis and the Lyapunov time on the horizontal axis. The Lyapunov time is defined by $\Lambda_{max}t$ where $\Lambda_{max}$ is the largest Lyapunov exponent computed from Eq.(\ref{eq:KS}). Predictions of this evolution by four different methods are shown in panels (B), (C), (D), and (E). The prediction in (B) is made by a single reservoir computer with 16,000 nodes, the prediction in (C) is made by 16 reservoir computers in parallel each with 4,000 nodes, the prediction in (D) is made by a hybrid with a single reservoir computer with 16,000 nodes(D), and the prediction in (E) is made by CHyPP with 16 reservoir computers in parallel each with 4,000 nodes and an imperfect model with second derivative $\epsilon=0.1$. The plots on the right (b), (c), (d), and (e) display the difference between the true dynamics (A) and the prediction using each of the corresponding techniques. The vertical black line marks the valid prediction time. Each technique was trained using a 3,375 Lyapunov time sequence of training data.}
    \label{fig:predictions}
\end{figure}
\begin{figure}[h!]
    \centering
    \includegraphics[width = 0.8\textwidth]{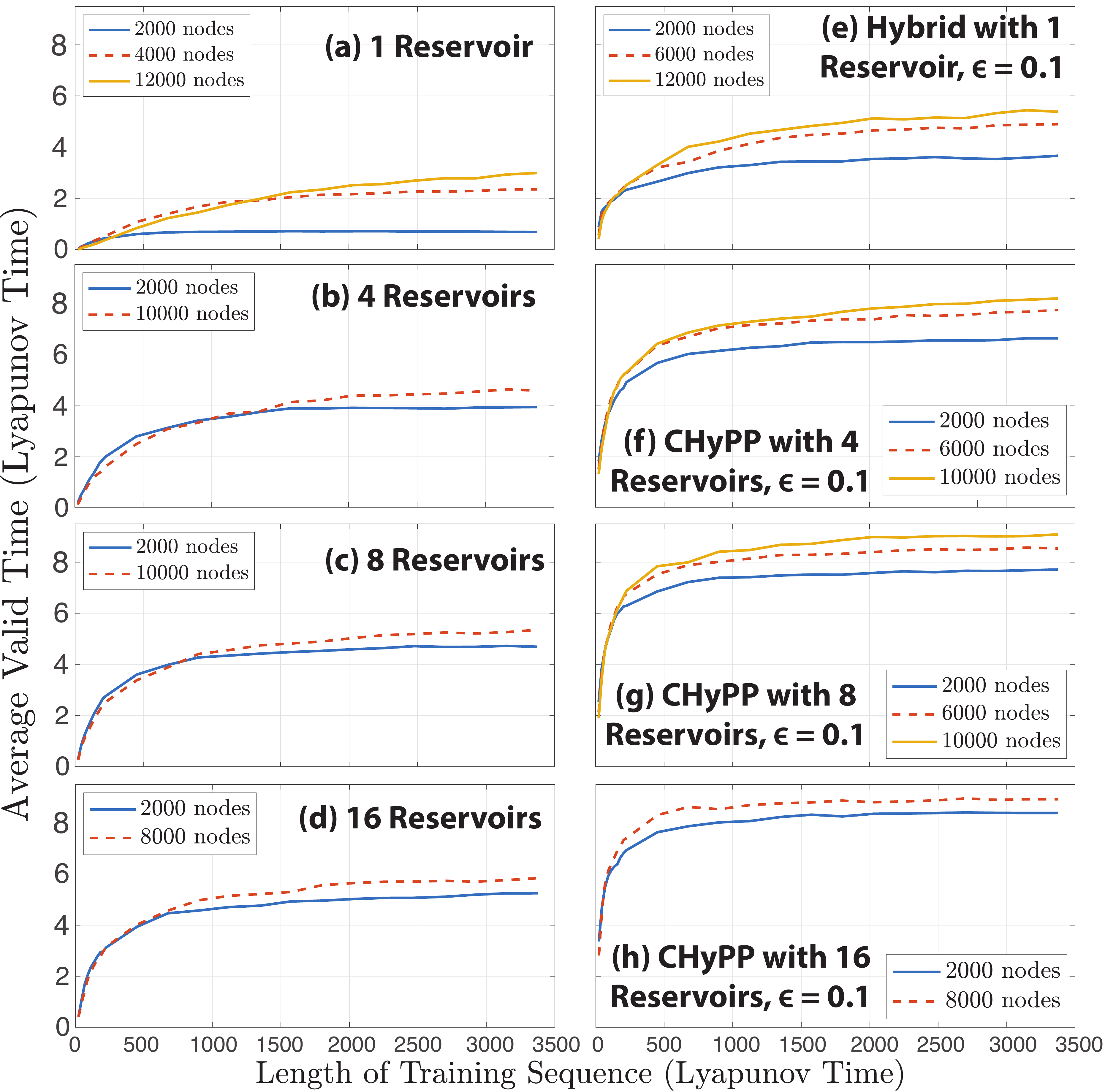}
    \caption{We plot the average valid time as a function of the duration of the training data, each in Lyapunov time, for the parallel reservoir(s)-only method using (a) 1 reservoir, (b) 4 reservoirs, (c) 8 reservoirs, and (d) 16 reservoirs. We plot the same quantity obtained from the parallel hybrid method using (e) 1 reservoir, (f) 4 reservoirs, (g) 8 reservoirs, and (h) 16 reservoirs. In each case, the imperfect knowledge-based predictor in the hybrid prediction had an error in the second derivative term of $\epsilon = 0.1$, resulting in an average valid time for the model-only prediction of 0.48 of a Lyapunov time. A knowledge-based predictor with no error begun using the most accurate initial condition from the parallel hybrid was able to predict for an average of 11 Lyapunov times. The legend in each plot indicates the number of nodes used in each reservoir.}
    \label{fig:vstrain}
\end{figure}

The prediction results shown in Figs.~\ref{fig:predictions} and~\ref{fig:vstrain} use the parameters contained in Table ~\ref{tab:params1}. In addition, the number of reservoirs and the length of training data are specified in the figure captions. The imperfect model has an error of $\epsilon = 0.1$ and a small valid prediction time of only $0.48$ Lyapunov times. Each plotted valid time is averaged over 100 predictions that are generated using the same set of reservoirs with the same training data sequence but that are synchronized to different initial conditions. Figure~\ref{fig:predictions} displays a set of example predictions of one of these test time series. Figures~\ref{fig:predictions} and~\ref{fig:vstrain} both demonstrate that CHyPP yields significantly longer valid predictions than the parallel reservoir-only method, which (for our choice of $\epsilon$ and reservoir size) produces longer valid predictions than the imperfect model alone. Both the parallel hybrid and parallel reservoir-only predictions outperform the imperfect model-only approach. From Fig.~\ref{fig:vstrain}, we also observe that the CHyPP saturates, or reaches a valid prediction time that increases only negligibly with the addition of more training data, at a much shorter length of training data than the parallel reservoir-only. As a result of this, the length of training data before which we would not benefit from increasing the size of the reservoir is also much shorter in the CHyPP prediction. For example, consider the plots in (b) and (f) in Fig.~\ref{fig:vstrain}, where each prediction uses 4 reservoirs. If we have a 500 Lyapunov time length of training data, we would not benefit from increasing the number of nodes per reservoir above 2000 in the reservoirs-only prediction, but could obtain a significant performance improvement if we did so using CHyPP. CHyPP exhibits its most impressive performance for very short lengths of training data. With only 22.5 Lyapunov times worth of training data, the parallel reservoir-only method is able to predict for only 0.44 Lyapunov times, whereas CHyPP with 16 reservoirs of 2000 nodes each predicts for 3.35 Lyapunov times on average, matching the saturated single reservoir-only prediction that is trained using 150 times more training data.
\subsection{Prediction Quality dependence on local overlap length}\label{sec:locality}
\begin{figure}[h]
    \centering
    \includegraphics[width = \textwidth]{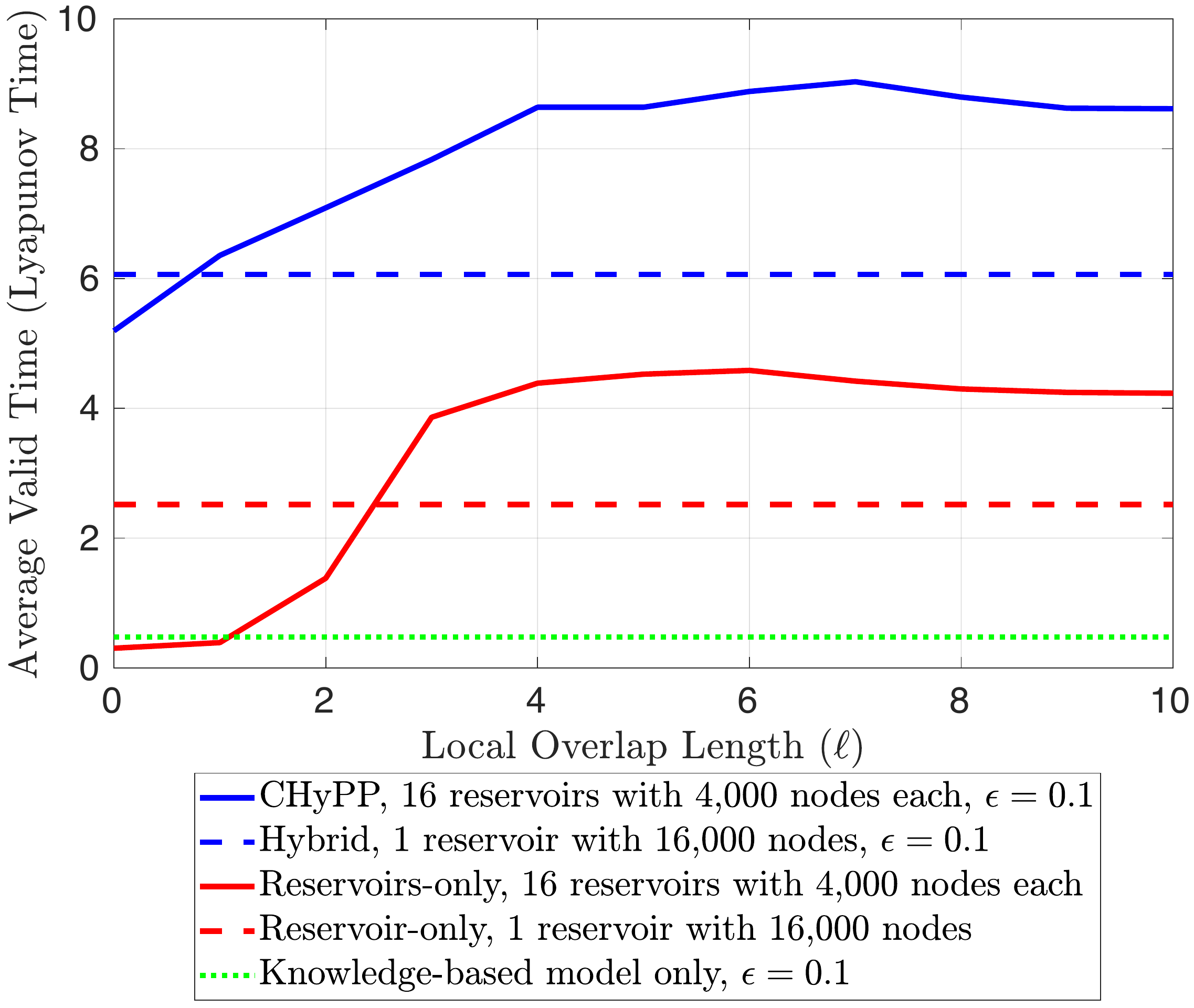}
    \caption{We plot the valid time for the parallel and non-parallel hybrid and reservoir(s)-only prediction as a function of the local overlap length $\ell$.}
    \label{fig:vslocality}
\end{figure}

In this section, we investigate the dependence of CHyPP performance on the local overlap length $\ell$. Figure~\ref{fig:vslocality} shows that when the local overlap length $\ell$ is zero or close to zero, the parallel reservoir-only prediction valid time (solid red line) is very poor, predicting for only around 0.5 Lyapunov times. 
CHyPP, however, is still able to obtain good predictions with very little local overlap length. For CHyPP, this indicates that it is able to utilize the inaccurate model prediction to infer the propagation of dynamical influences between the adjacent prediction regions. We note that neither CHyPP nor the 16 parallel reservoirs-only prediction are able to improve upon a corresponding single, large reservoir hybrid prediction or corresponding single large reservoir-only prediction when the local overlap length is near 0. Regarding this comparison, however, it is important to recognize that in systems larger than the one we have used to test our method here, use of a single reservoir prediction is not possible because the reservoir size necessary for prediction becomes infeasible, as discussed in Sec.~\ref{sec:reservoir}. This result is nevertheless very important for large-scale implementations of the proposed method. In realistic systems with 3 spatial dimensions, a small increase in the local overlap length $\ell$ can lead to increasingly large memory requirements and increasingly large amounts of information that must be communicated between reservoirs at each prediction iteration. Being able to obtain a good prediction with less local overlap length when the propagating dynamics is adequately explained by the imperfect model is thus another advantage of CHyPP.

\section{Test Results on a Multiscale System Prediction and the Problem of Subgrid-Scale Closure}\label{sec:multiscale}
A common difficulty in numerical modeling arises because many physical processes involve dynamics on multiple scales. As a result, fundamentally crucial subgrid scale dynamics is often only crudely captured in an ad-hoc manner. The formulation of subgrid scale models by analytical techniques has been extensively studied (e.g., see Ref.~\cite{meneveau_scale-invariance_2000}) and is sometimes referred to as ``subgrid-scale closure''. In this section, we show that our CHyPP methodology provides a very effective data-based (as opposed to analysis-based) approach to subgrid-scale closure. 

In particular, we test our CHyPP prediction method on a dynamical system with multiple spatial and temporal scales. The particular system we choose to predict is the multiscale ``toy'' atmospheric model formulated by E.N. Lorenz in his 2005 paper~\cite{lorenz2005}. We refer to this model as Lorenz Model III. This model is a smoothed extension of Lorenz's original ``toy'' atmospheric model described in his 1996 paper~\cite{lorenz1996predictability} (hereafter referred to as Lorenz Model I), with the addition of small scale dynamical activity. Lorenz Model III describes the evolution of a single atmospheric variable, $Z$, on a one-dimensional grid with $N$ grid points and periodic boundary conditions, representing a single latitude. The value of $Z$ at each grid point, $Z_n$, evolves according to the following equation,
\begin{align}\label{eq:lorenz3}
    dZ_n/dt = &[X, X]_{K, n}+b^{2}[Y, Y]_{1, n}+c[Y, X]_{1, n}-X_{n} -b Y_{n}+F.
\end{align}
In Eq.(\ref{eq:lorenz3}), $X$ is a smoothed version of $Z$, and $Y$ is the difference between $Z$ and $X$,
\begin{align}
        X_{n}=&\sum_{i=-I}^{I}(\alpha-\beta|i|) Z_{n+i},\quad Y_{n}=Z_{n}-X_{n},\\
        \alpha=& \left(3 I^{2}+3\right) /\left(2 I^{3}+4 I\right),\quad \beta=\left(2 I^{2}+1\right) /\left(I^{4}+2 I^{2}\right).
\end{align}
Here, $I$ denotes the smoothing distance. Thus, $X$ describes the large-spatial-scale, long-time-scale wave component of $Z$, while $Y$ describes the small-spatial-scale, short-time-scale wave component. $[V,W]_{K,n}$ indicates a coupling between the variables $V$ and $W$ [i.e., interaction within scales ($V=W=X$ or $Y$) or between scales for ($V = X$, $W = Y$)]. For $K$ odd, this coupling takes the form,
\begin{align}
    [V, W]_{K, n}= \sum_{j=-J}^J\textrm{$'$} \sum_{i=-J}^J\textrm{$'$}(-V_{n-2 K-i}W_{n-K-j}+V_{n-K+j-i} W_{n+K+j}) / K^2.
\end{align}
Here, $J = (K-1)/2$ and $\sum \textrm{$'$}$ denotes a modified summation where the first and last summands are divided by 2. For $K$ even, $J = K/2$ and each $\sum '$ becomes a standard summation $\sum$. In Eq.(\ref{eq:lorenz3}), the parameters $K$, $b$, $c$, and $F$ describe the coupling distance of the system's large scale dynamics, the increase in small scale oscillation rapidity and decrease in amplitude (relative to the large scale dynamics), the degree of interaction between the large and small scale dynamics, and the overall forcing in the system, respectively. We note that when $b=c=0$ and $K=1$, this model reduces to the Lorenz Model I.

For the purposes of testing our CHyPP method, we also introduce another model formulated by Lorenz, which we will refer to as Lorenz Model II~\cite{lorenz2005}. This model is equivalent to Lorenz Model III with no distinct small scale wave component or smoothing present in the equation,
\begin{align}\label{eq:lorenz2}
    d Z_{n} / d t=[Z, Z]_{K, n}-Z_{n}+F.
\end{align}
Important for our tests, Lorenz notes that, for constant $F$, the dominant wavenumber in Lorenz Model II depends only on the ratio $N/K$. 

We test our CHyPP method by using it to predict the dynamics generated from Lorenz Model III using Lorenz Model II as our imperfect knowledge-based predictor. As we discussed in Sec.~\ref{sec:intro}, knowledge-based models used to predict physical multiscale systems (i.e., the weather) often use simplified representations of subgrid scale dynamics. To replicate this type of imperfection in the knowledge-based model in CHyPP, we realize the Lorenz Model II dynamics over an equal or fewer number of grid points $N_{\textrm{Model II}}$ while using the same $N/K$ ratio as used to generate the true Lorenz Model III dynamics (i.e., $N_{\textrm{Model III}}/K_{\textrm{Model III}} = N_{\textrm{Model II}}/K_{\textrm{Model II}}$). Our imperfect model thus incorrectly represents the effect of small scale dynamics that, when $N_{\textrm{Model III}} > N_{\textrm{Model II}}$, are also subgrid.

\begin{table}[h!]
\centering
\begin{tabular}{|c|c||c|c|}
    \hline
    $\langle d \rangle$ (average in-degree) & 3 & $\rho$ (spectral radius) & 0.6\\
    \hline
    $\ell$ (local overlap length) & N/12 & $\sigma$ (input coupling strength) & 1\\
    \hline
    $\Delta t$ (Prediction time step) & 0.005 & $\beta$ (regularization) & $10^{-4}$ \\
    \hline
    $T_S$ (synchronization time) & 0.5& &\\ 
    \hline
\end{tabular}
\caption{Hyperparameters}
\label{tab:params2}
\end{table}

For the following results in this section, we use the parameters in Table~\ref{tab:params2}. In addition, the value of $s$ used in each of the reservoir computing-based prediction methods has been selected for each method to maximize the valid prediction time. 
\begin{figure}[h]
    \centering
    \includegraphics[width = \textwidth]{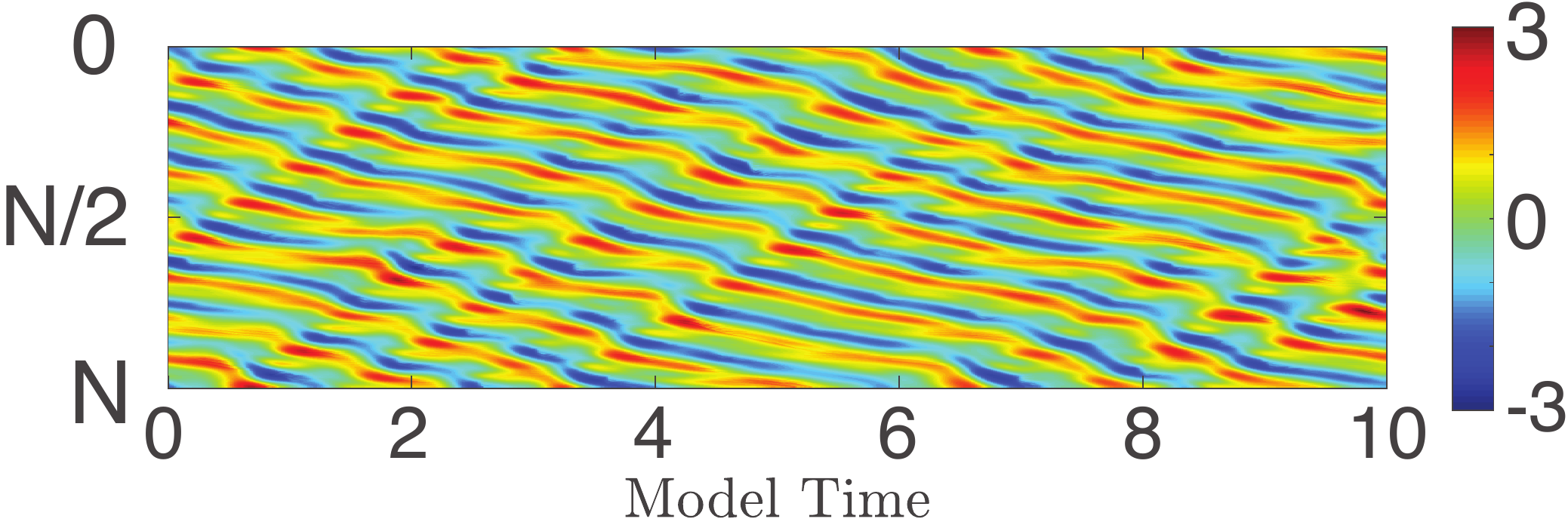}
    \caption{Plot of the true dynamics from Lorenz Model III using the parameters $N = 960$, $K = 32$, $I = 12$, $F = 15$, $b = 10$, and $c = 2.5$. The dynamics have been normalized to set the mean value to 0 and variance to 1. We plot the value of $Z$ color-coded (color bar on the right), the spatial grid point along the vertical axis, and the model time on the horizontal axis.}
    \label{fig:model3_truth}
\end{figure}
\begin{figure}[h]
    \centering
    \includegraphics[width = \textwidth]{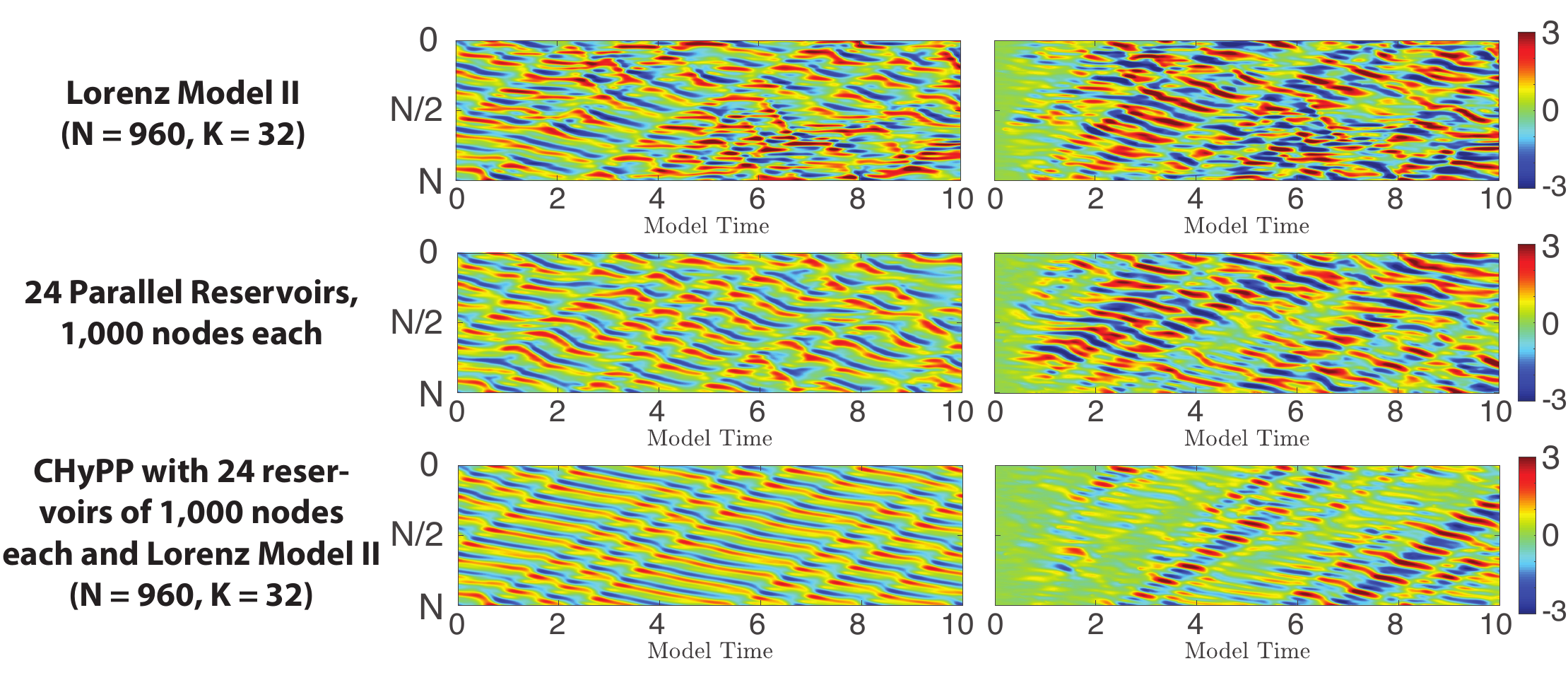}
    \caption{We plot the predictions of the normalized true dynamics from Lorenz Model III (displayed in Fig.~\ref{fig:model3_truth}) where we have measured every spatial grid point of Model III during training. We plot the spatial grid point along the vertical axis and the model time on the horizontal axis. The plots on the left show the predictions made using each of the specified methods, while the plots on the right show the difference between the true dynamics and the prediction (i.e., the prediction error). The parallel reservoir-only prediction used its optimized value of $s = 0.1$, while CHyPP used its optimized value of $s = 0.085$.}
    \label{fig:model3Nskip1pred}
\end{figure}
\begin{figure}[h]
    \centering
    \includegraphics[width = \textwidth]{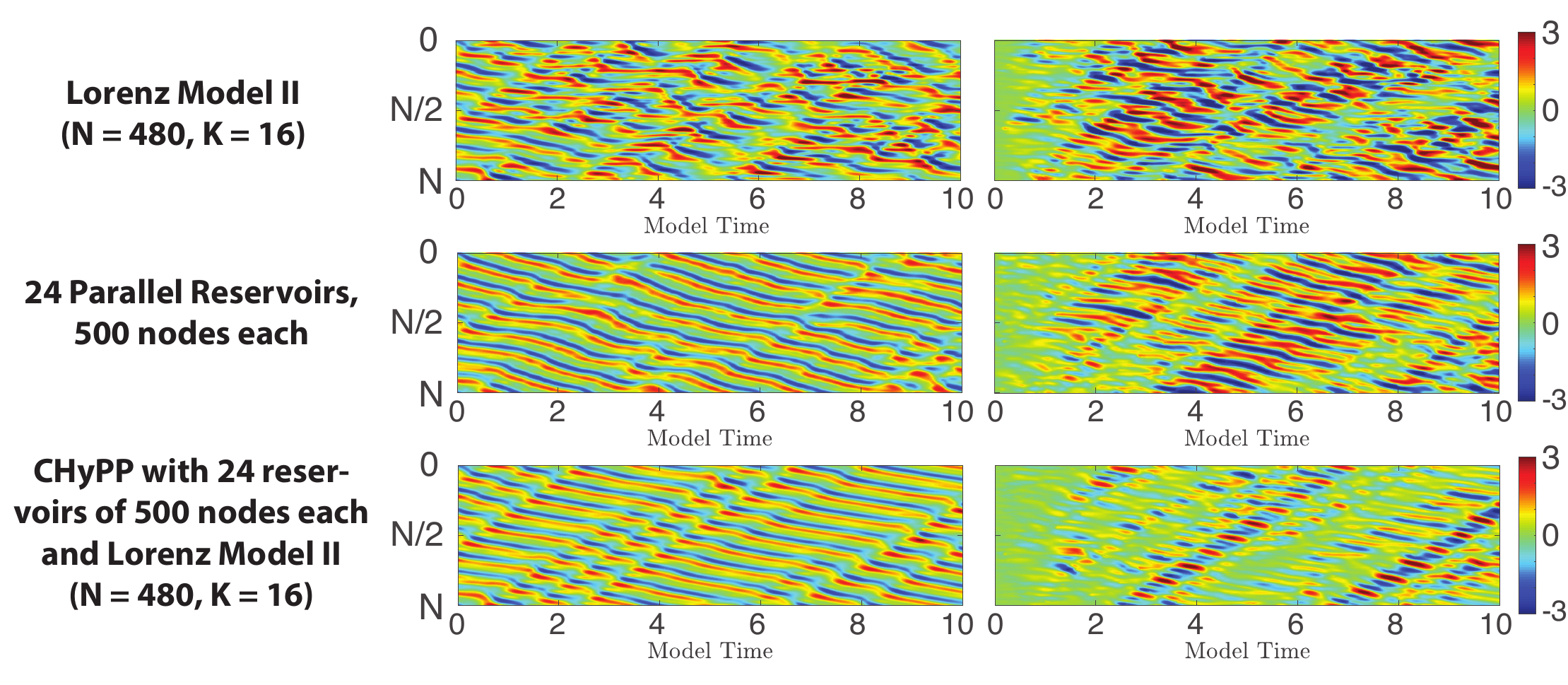}
    \caption{We plot the predictions of the normalized true dynamics from Lorenz Model III (displayed in Fig.~\ref{fig:model3_truth}), where, corresponding to the Model II grid, we have measured \textit{only every second} spatial grid point of Model III during training. We plot the spatial grid point along the vertical axis and the model time on the horizontal axis. The plots on the left show the predictions made using each of the specified methods, while the plots on the right show the difference between the true dynamics and the prediction (i.e., the prediction error). The parallel reservoir-only prediction used its optimized value of $s = 0.25$, while CHyPP used its optimized value of $s = 0.08$.}
    \label{fig:model3Nskip2pred}
\end{figure}
\begin{figure}[h]
    \centering
    \includegraphics[width = \textwidth]{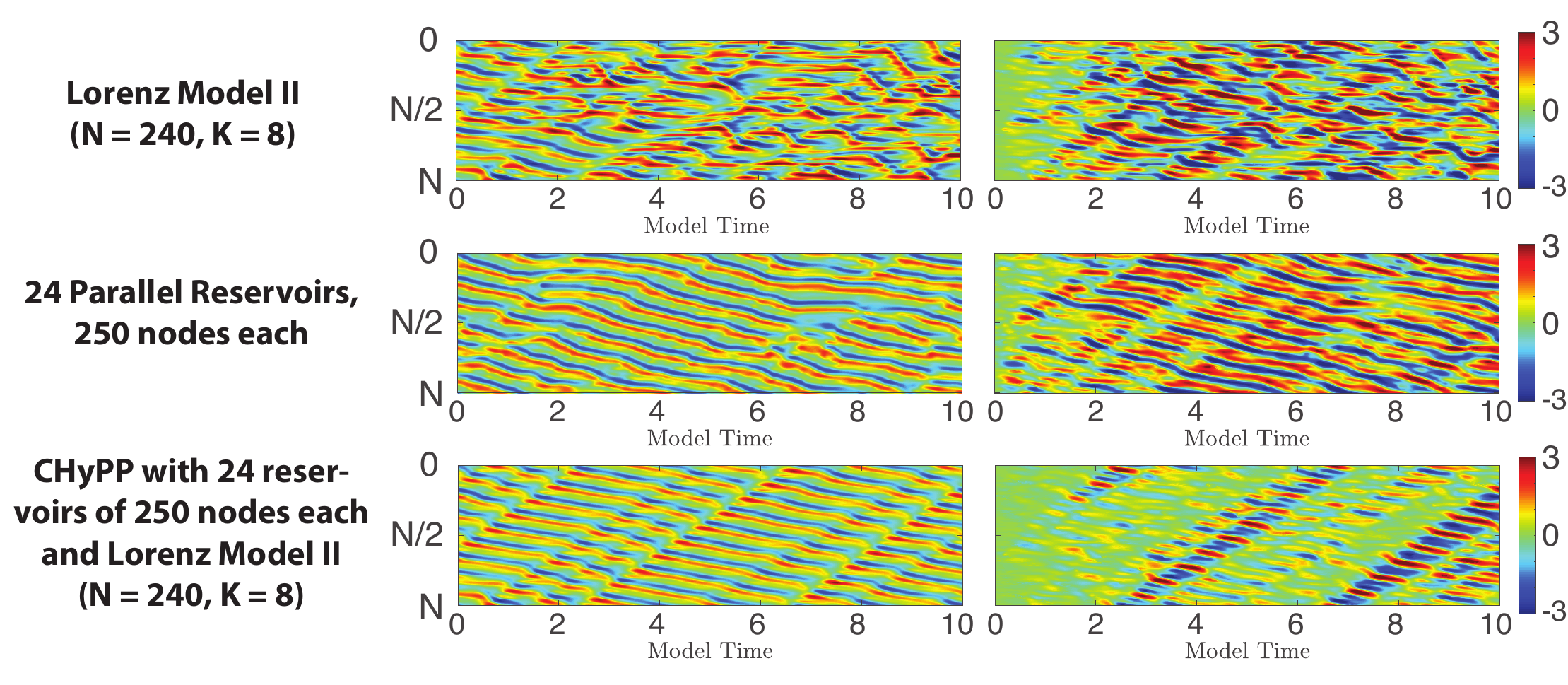}
    \caption{We plot the predictions of the normalized true dynamics from Lorenz Model III (displayed in Fig.~\ref{fig:model3_truth}), where, corresponding to the Model II grid, we have measured \textit{only every fourth} spatial grid point of Model III during training. We plot the spatial grid point along the vertical axis and the model time on the horizontal axis. The plots on the left show the predictions made using each of the specified methods, while the plots on the right show the difference between the true dynamics and the prediction (i.e., the prediction error). The parallel reservoir-only prediction used its optimized value of $s = 0.25$, while CHyPP used its optimized value of $s = 0.017$.}
    \label{fig:model3Nskip4pred}
\end{figure}
\begin{figure}[h]
    \centering
    \includegraphics[width = \textwidth]{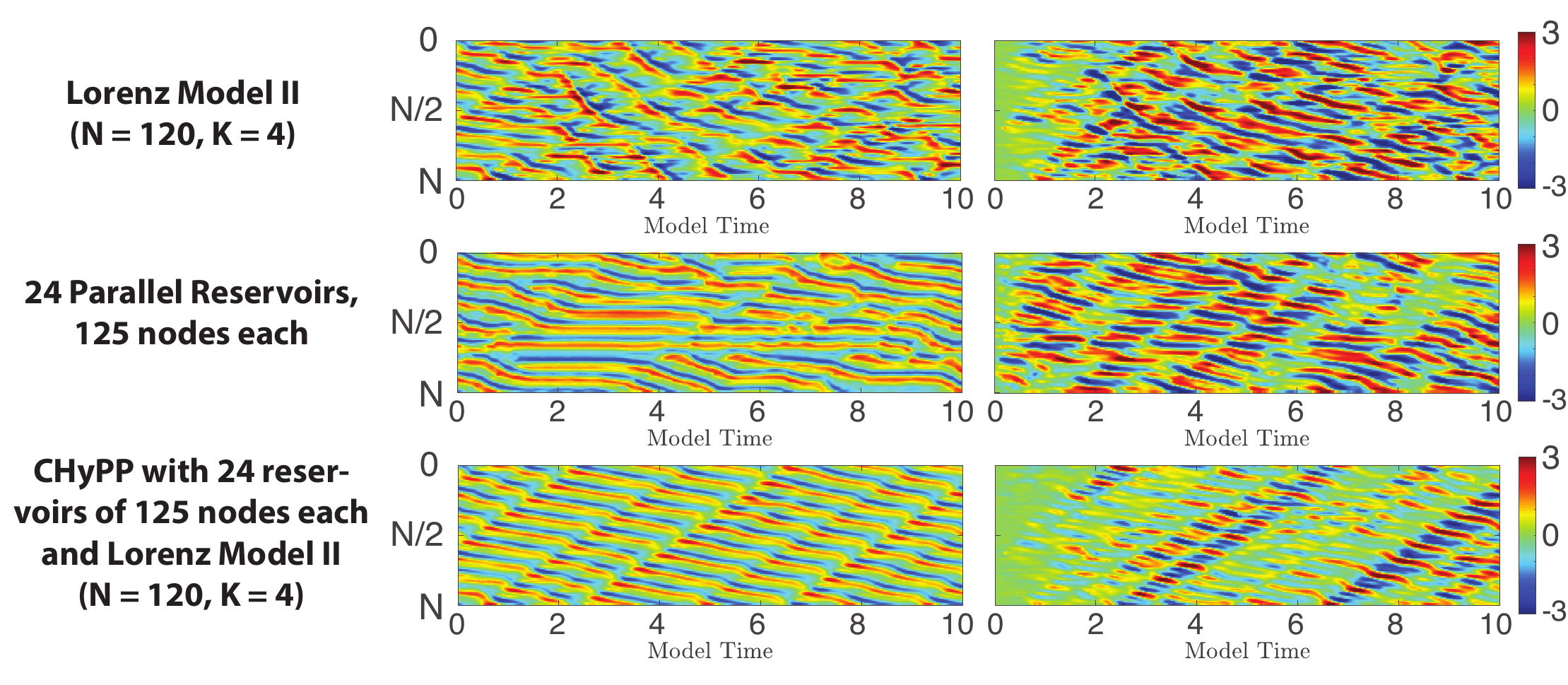}
    \caption{We plot the predictions of the normalized true dynamics from Lorenz Model III (displayed in Fig.~\ref{fig:model3_truth}), where, corresponding to the Model II grid, we have measured \textit{only every eighth} spatial grid point of Model III during training. We plot the spatial grid point along the vertical axis and the model time on the horizontal axis. The plots on the left show the predictions made using each of the specified methods, while the plots on the right show the difference between the true dynamics and the prediction (i.e., the prediction error). The parallel reservoir-only prediction used its optimized value of $s = 0.25$, while CHyPP used its optimized value of $s = 0.0155$.}
    \label{fig:model3Nskip8pred}
\end{figure}
Figure~\ref{fig:model3_truth} shows a solution of Lorenz Model III where the value of $Z$ is color-coded. The spatial variable is plotted vertically, time is plotted horizontally, and the model parameters are given in the caption. As seen in Fig.~\ref{fig:model3_truth}, there is a wave-like motion with a dominant wavenumber of $\approx 7$ (7 oscillations along the vertical periodicity length). This corresponds to Lorenz's design of the model to mimic atmospheric dynamics, which has a predominant wavenumber for Rossby waves of this order as one goes around a mid-latitude circle. Figures~\ref{fig:model3Nskip1pred}-\ref{fig:model3Nskip8pred} show predictions of the true Lorenz Model III dynamics (in Fig.~\ref{fig:model3_truth}) for grid resolutions of varying coarseness. In particular, while the truth (Fig.~\ref{fig:model3_truth}) is obtained from Model III with $N=960$ grid points, the number of Model II grid points is $N=960$, $480$, $240$, and $120$ for Figs.~\ref{fig:model3Nskip1pred},~\ref{fig:model3Nskip2pred},~\ref{fig:model3Nskip4pred}, and~\ref{fig:model3Nskip8pred} respectively. Also note that the measurements are taken from the Model III result only at the Model II grid points. For both the parallel reservoir-only predictions and CHyPP, we fix the ratio of the number of nodes per reservoir to the number of grid points each reservoir predicts to be $D_r/Q = 25$. 

\begin{figure}[h]
    \centering
    \includegraphics[width = \textwidth]{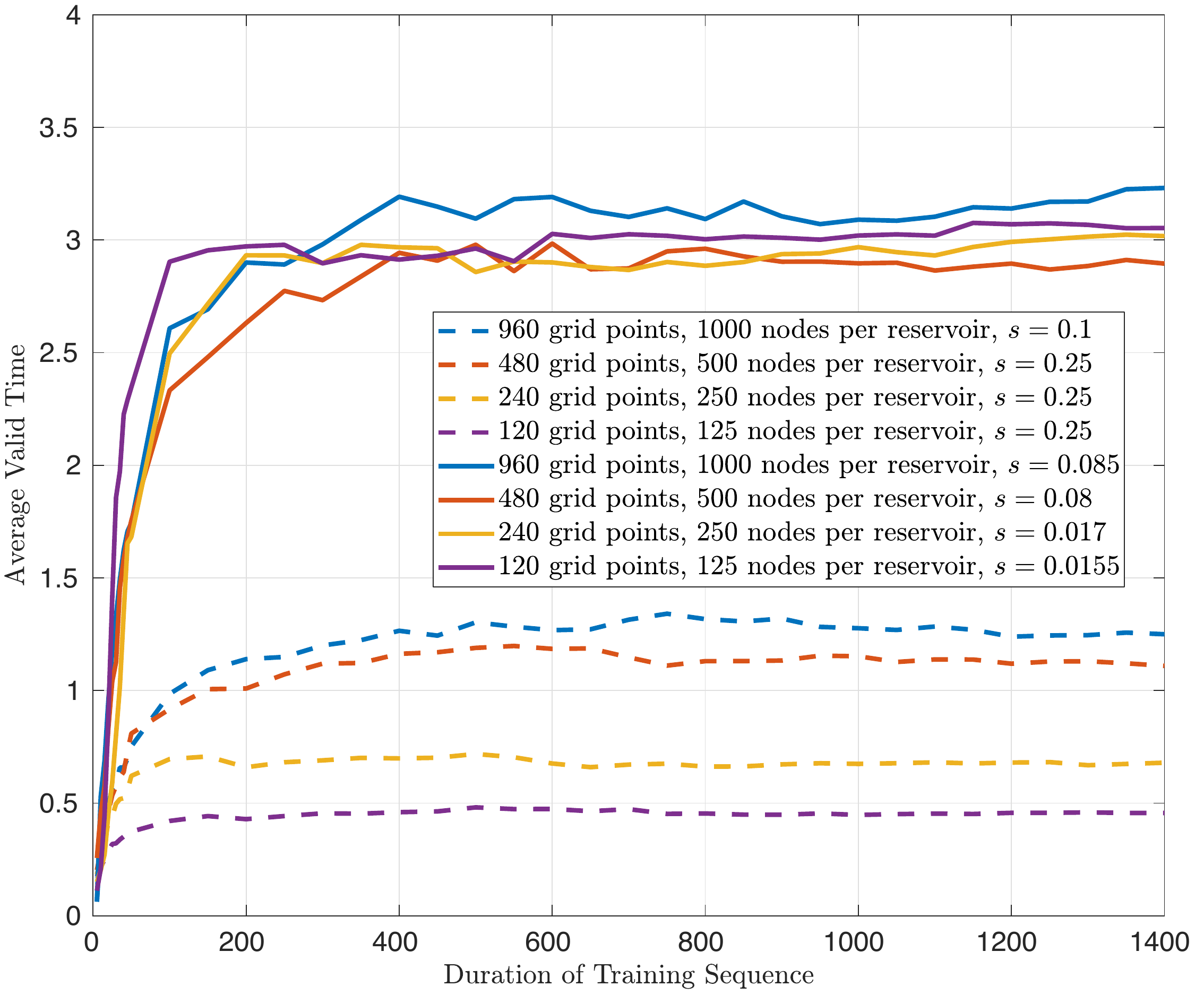}
    \caption{We plot the average valid time for each prediction as a function of the length of training data used. Dashed lines display the results of parallel reservoir-only predictions, whereas solid lines display results from CHyPP where the knowledge-based model is Lorenz Model II with $N =$ Number of grid points, $K = N/30$, and $F = 15$. Each prediction method uses 24 reservoirs.}
    \label{fig:model3vstrain}
\end{figure}

We find that CHyPP significantly outperforms each of its component methods, and that this is true for all of the grid resolutions we tested. We note that, unlike in the Model II and parallel reservoir-only predictions, the prediction error in CHyPP seems to appear locally in a small region and manifests as slanted streaks in the error plots in Figs.~\ref{fig:model3Nskip1pred}-\ref{fig:model3Nskip8pred}. We have verified that the slant of these streaks corresponds to the group velocity of the dominant wave motion (wavenumber $\approx 7$). For all predictions made, we again calculate a valid time of prediction; however, since in this case early error growth using CHyPP is local in space and affects only a small part of the prediction domain, we choose to use a higher valid time error threshold of 0.85 (approximately 60\% of the error saturation value) so that the valid time metric reflects when error is present in the entire prediction domain. Figure~\ref{fig:model3vstrain} shows the valid time averaged over 100 predictions from different initial conditions versus training time for different grid resolutions and reservoir sizes. The dashed lines are parallel reservoir-only predictions, while the solid lines are CHyPP predictions. In this figure, the reservoir sizes are scales in proportion to the number of Model II grid points used. We find that, while the quality of the scaled parallel reservoir-only prediction degrades significantly as the grid resolution decreases, the quality of the scaled CHyPP prediction degrades much more slightly from an average valid time of $3.23$ at full resolution to $3.05$ at $1/8$ resolution.
The Model II valid time is essentially constant at roughly $0.95$, corresponding to the fact that all of the grid spacings tested are well below the characteristic Model II spatial scale. We see from Fig.~\ref{fig:model3vstrain} that the parallel reservoir-only prediction and CHyPP prediction appear to reach valid time saturation at about the same training data length for each grid resolution.
\section{Conclusion}\label{sec:conclusion}
In this paper, we address the general goal of utilizing machine learning to enable expanded capability in the forecasting of a large, complex, spatiotemporally chaotic systems for which an imperfect knowledge-based model exists. Some typical common sources of imperfection in such a knowledge-based model are unresolved subgrid scale processes and lack of first principles knowledge or computational ability for modeling some necessary aspect or aspects of the physics. The hope is that these ``imperfections" can be compensated for by use of measured time series data and machine learning. The two main foreseeable difficulties in realizing this hope are how to effectively combine the machine learning component with the knowledge-based component in such a way that they mutually enhance each other, and how to promote feasible scaling of the machine learning requirements with respect to its computational cost and necessary amount of training data. Note that addressing the first of these issues necessarily lessens the difficulty of dealing with the second issue, since good use of any valid scientific knowledge of the system being forecast can potentially reduce the amount of learning required from the machine learning component.

To address these two issues, we propose a methodology (CHyPP) that combines two previously proposed techniques: (i) a hybrid utilization of an imperfect knowledge-based model with a single machine learning component~\cite{hybridreservoir2018,wan2018data}, and (ii) a parallel machine learning scheme using many spatially distributed machine learning devices~\cite{parallelreservoir2018}. We note that (ii) applies to large spatial systems with the common attribute of what we have called `local short-time causal interactions'. Numerical tests of our proposed combination of (i) and (ii) are presented in Secs.~\ref{sec:results} and~\ref{sec:multiscale} and demonstrate good quality prediction that is scalable with size (Figs.~\ref{fig:rmsescalability} and~\ref{fig:predictions}), reduced required length of training data (Fig.~\ref{fig:vstrain}), and ability to compensate for unresolved subgrid-scale processes (Figs.~\ref{fig:model3_truth}-\ref{fig:model3vstrain}).

In this paper, our proof-of-principle problems have been one-dimensional in space with relatively simple mathematical formulations. We note, however, that the CHyPP methodology is readily applicable to higher dimensions and more complicated situations. For example, we are currently in the process of applying CHyPP to global atmospheric weather forecasting for which the atmospheric state is spatially three-dimensional and the system is strongly inhomogeneous, e.g., due to the presence of complex geographic features (including continents, mountains, and oceans), as well as the latitudinal variation of solar heating.

In conclusion, we have shown that data-assisted forecasting via parallel reservoir computing and an imperfect knowledge-based model can significantly improve prediction of a large spatiotemporally chaotic system over existing methods. This method is scalable to even larger systems, requires significantly less training data than previous methods to obtain high quality predictions, is able to effectively utilize a knowledge-based forecast of information propagation between local regions to improve prediction quality, and effectively provides a means of using data to compensate for unresolved subgrid scale processes (playing a role akin to traditional analysis-based closure schemes).

\section*{Acknowledgements}
We thank Sarthak Chandra for his helpful discussion and comments. This work was supported by DARPA contract DARPA-PA-18-01(HR111890044).

\bibliography{hybrid_parallel_bibliography}

\end{document}